\documentclass{article}

\usepackage{arxiv}

\usepackage[utf8]{inputenc} 
\usepackage[T1]{fontenc}    
\usepackage{hyperref}       
\usepackage{url}            
\usepackage{booktabs}       
\usepackage{amsfonts}       
\usepackage{nicefrac}       
\usepackage{microtype}      
\usepackage{lipsum}		
\usepackage{graphicx}
\usepackage{natbib}
\usepackage{doi}
\usepackage{amsmath}

\usepackage{graphics}
\usepackage{caption}
\usepackage[caption=false,font=normalsize,labelfont=sf,textfont=sf]{subfig}
\usepackage{multirow}
\usepackage{subfig}
\usepackage{hyperref}

\title{Beyond Scores: A Modular RAG-Based System for Automatic Short Answer Scoring with Feedback}

\author{{\includegraphics[scale=0.06]{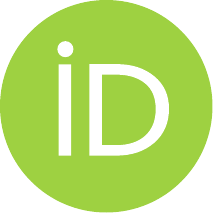}\hspace{1mm}Menna Fateen} \\
Graduate School of Information Science and Electrical Engineering\\
Kyushu University\\
Fukuoka, 8190395, Japan \\
\texttt{menna.fateen@m.ait.kyushu-u.ac.jp} \\
\And
{\includegraphics[scale=0.06]{orcid.pdf}\hspace{1mm}Bo Wang} \\
Graduate School of Information Science and Electrical Engineering\\
Kyushu University\\
Fukuoka, 8190395, Japan \\
\texttt{wangbo.rw@gmail.com} \\
\And
{\includegraphics[scale=0.06]{orcid.pdf}\hspace{1mm}Tsunenori Mine} \\
Faculty of Information Science and Electrical Engineering\\
Kyushu University\\
Fukuoka, 8190395, Japan \\
\texttt{mine@ait.kyushu-u.ac.jp } \\
}

\renewcommand{\shorttitle}{\textit{arXiv} Template}

\hypersetup{
pdftitle={A template for the arxiv style},
pdfsubject={q-bio.NC, q-bio.QM},
pdfauthor={David S.~Hippocampus, Elias D.~Striatum},
pdfkeywords={automatic short answer scoring, large language models, retrieval-augmented generation},
}

\begin{document}
\maketitle

\begin{abstract}
	Automatic short answer scoring (ASAS) helps reduce the grading burden on educators but often lacks detailed, explainable feedback. Existing methods in ASAS with feedback (ASAS-F) rely on fine-tuning language models with limited datasets, which is resource-intensive and struggles to generalize across contexts. Recent approaches using large language models (LLMs) have focused on scoring without extensive fine-tuning. However, they often rely heavily on prompt engineering and either fail to generate elaborated feedback or do not adequately evaluate it. In this paper, we propose a modular retrieval augmented generation based ASAS-F system that scores answers and generates feedback in strict zero-shot and few-shot learning scenarios. We design our system to be adaptable to various educational tasks without extensive prompt engineering using an automatic prompt generation framework. Results show an improvement in scoring accuracy by 9\% on unseen questions compared to fine-tuning, offering a scalable and cost-effective solution.

\end{abstract}


\section{Introduction}
\label{sec:introduction}

Feedback plays a critical role in student learning. Extensive research demonstrates that detailed and timely feedback significantly improves student performance by guiding learning and correcting misunderstandings \cite{shute2008focus,hattie2007power}. However, providing such elaborated feedback, especially for short answer questions, is a time-consuming and labor-intensive task for educators, particularly when managing large numbers of students. To address this, automatic short answer scoring (ASAS) systems have been developed to quickly assess student responses \cite{riordan2017investigating,ramachandran2015identifying}. Despite their utility, ASAS systems primarily focus on scoring correctness, offering holistic scores without the detailed feedback students need to understand their mistakes \cite{burrows2015eras}.

Current ASAS systems are typically trained on classification or regression tasks, where the objective is to assign a score to a student's answer. While effective in evaluating answers, these systems lack the explainability needed for two key purposes: (1) helping students understand the reasoning behind their errors and the feedback given and (2) fostering trust in the scoring system \cite{winstone2017supporting,liu2018towards,khosravi2022explainable}. Without this granularity, students miss opportunities to improve, and educators may be reluctant to adopt automated grading systems.

In response to these limitations, recent research has focused on developing ASAS systems with feedback \textbf{(ASAS-F)} that not only score student answers but also generate detailed feedback \cite{li2023distilling,schneider2023towards,filighera2022your}. These systems aim to provide students with actionable feedback that explains the reasoning behind their scores, helping them identify and correct their mistakes. Existing ASAS-F systems rely on resource-intensive fine-tuning with limited datasets, specific tasks \cite{dzikovska2014beetle,keuning2018systematic} or complex prompt engineering, all of which may not generalize well across contexts. Moreover, evaluating the quality of the textual feedback is usually only done using traditional statistical metrics, which do not capture the main aspects of quality, such as accuracy and clarity.

To overcome these challenges, we introduce a novel ASAS-F system that leverages large language models (LLMs) within a Retrieval-Augmented Generation (RAG) \cite{lewis2020retrieval} framework. Inspired by similarity-based scoring \cite{bexte2023similarity}, our method retrieves the most similar answers from a short answer scoring feedback dataset using the ColBERT retrieval \cite{khattab2020colbert} model. These retrieved answers serve as few-shot examples, allowing LLMs to generate both accurate scores and detailed feedback.

Our study explores the following research questions:
\begin{itemize}
  \item \textbf{RQ1:} How does the performance of our modular ASAS-F system compare to state-of-the-art models in automatic short answer scoring?
  \item \textbf{RQ2:} When labeled training data is available, how can we optimize prompts and few-shot examples to improve our ASAS-F performance in an efficient way?
  \item \textbf{RQ3:} How accurate and clear is the feedback generated by our ASAS-F system?
\end{itemize}

Our experiments on the research questions demonstrate that the proposed approach significantly minimizes the need for extensive fine-tuning, resulting in a computationally efficient solution that maintains high accuracy while delivering clear, accurate feedback. The system is designed in a modular fashion, allowing it to adapt easily to various educational tasks without the need for extensive prompt engineering. By reducing reliance on large curated datasets, our method generates meaningful feedback with minimal adaptation, offering a scalable and cost-effective solution for educational applications. We share the code, model outputs and their evaluations from our experiments on Github to promote transparency and reproducibility \footnote{Link available upon acceptance}.

To summarize, the key contributions of our work are:

\begin{itemize}
  \item We propose a novel ASAS-F system that scores short answers and generates detailed feedback using LLMs and RAG.
  \item We propose a modular approach that minimizes the need for prompt engineering, making the system adaptable to diverse tasks 
  \item We propose a RAG-based approach that eliminates the need for computationally expensive fine-tuning and large datasets, making the solution scalable and efficient.
\end{itemize}

\section{Related Work}

\subsection{Automatic Short Answer Scoring}

Automatic scoring has been studied extensively in the field of natural language processing (NLP) and educational technology \cite{burrows2015eras, xi2010automated,balfour2013assessing,ke2019automated}. ASAS systems aim to grade students’ short answers automatically, offering immediate feedback, reducing teachers’ workload, and streamlining the grading process.

Traditional ASAS systems are often rule-based and rely on handcrafted features and heuristics to score short answers. While these systems are effective and have shown promising results on benchmark datasets, they require significant manual effort to develop \cite{ramachandran2015identifying,kumar2019get}. With the rise of deep learning, neural network-based and Transformer-based ASAS systems have been introduced, which automatically learn features from data and outperform traditional rule-based systems \cite{riordan2017investigating,sung2019improving}.

Among the newer methods, instance-based approaches have gained traction. These approaches involve using pretrained models to map  student responses directly to scores by pooling token representations across model layers \cite{steimel2020towards}. These methods capture the context and semantics of input text, enabling them to outperform traditional rule-based systems. However, they often require large amounts of labeled data to achieve high performance and struggle in few-shot settings where data is limited.  Comparisons between instance-based and similarity-based approaches have shown the latter to be more effective in zero-shot settings \cite{bexte2023similarity,wang2023optimizing, fateen-mina-2023-context}. Additionally, recent advancements in dense information retrieval have proven effective in automatic essay scoring tasks \cite{albatarni2024graded}. Building on these insights, our work introduces a similarity-based approach using retrieval-augmented generation to enhance the performance of ASAS-F systems.

\subsection{ASAS Using Generative Models}

Despite the success of neural network-based ASAS systems, many of the proposed models are computationally expensive and require large amounts of data to achieve good performance. In many real-world scenarios, labeled data for training ASAS systems is scarce, making it challenging to train these models.

Few studies have explored the use of large language models (LLMs) for ASAS tasks. One study investigated the usage of LLMs for ASAS, focusing on zero-shot and few-shot settings across three diverse datasets \cite{chamieh2024llms}. The results showed that LLMs were able to achieve strong performance on tasks involving general knowledge questions but struggled with questions that required domain-specific knowledge or complicated reasoning.

Another study explored the use of proprietary LLMs, namely GPT3.5 and GPT4, for ASAS tasks \cite{chang2024automatic}, \cite{achiam2023gpt}. The findings indicated that while these models showed promise, their performance was negatively correlated with the length of student answers.

In a recent study, researchers examined the use of ChatGPT for holistic essay scoring and compared its performance to human raters \cite{tate2024can}. The findings showed that ChatGPT’s scoring was not statistically significantly different from human ratings and demonstrated substantial agreement. However, this study highlighted the need for further research particularly regarding improving scoring precision.

Additionally, research assessed the possibility of using ChatGPT (GPT-3.5) for automated grading \cite{schneider2023towards}. The results showed a low correlation between the scores given by ChatGPT and human graders, with ChatGPT tending to give middle scores more frequently despite strong variation in student answers. Moreover, the study found that minor changes in answers could lead to significant differences in the scores given by ChatGPT.

\subsection{ASAS with Feedback}

While scores provide a general overview of the quality of the answer, they lack the granularity and explainability that students need to improve their answers. Elaborated feedback would also increase the trustworthiness of the scoring system.

The Short Answer Feedback (SAF) dataset \cite{filighera2022your} includes short answer questions in communication network topics, each with a reference answer, given score, label, and content-focused elaborated feedback. The SAF dataset provides a benchmark for evaluating ASAS systems that generate feedback, enabling the development of more informative and actionable systems. It also established a baseline for ASAS systems using the T5 model fine-tuned on the SAF dataset, demonstrating the feasibility of generating feedback for short answers.

A framework was proposed \cite{li2023distilling} that distils the rationale generation capability of ChatGPT by designing several prompt templates to generate feedback from ChatGPT, then fine-tuning a smaller language model on the refined generated outputs. The study showed that the fine-tuned model outperformed the original ChatGPT model on QWK scores.

Further research explored four different approaches for generating counterfactual feedback for short answers \cite{filighera2022towards}. While some of the modified feedback could be graded as correct by automatic scoring systems, a domain expert deemed them incorrect.

Finally, one study explored the usage of LLMs for automated essay scoring (AES) with rationale \cite{yancey2023rating}, though the primary focus in ASAS is on content quality rather than writing style and structure. To the best of our knowledge, there is limited research evaluating LLMs for ASAS-F tasks, particularly in zero-shot and few-shot settings. Our work aims to bridge this gap by exploring the potential of LLMs for scoring short answers and generating feedback.

\section{Methodology}

In this section, we present our approach for building the ASAS-F system using LLMs and outline the methodologies proposed for different scenarios, including zero-shot ASAS-F,  few-shot ASAS-F with automatic optimization and few-shot ASAS-F using RAG.

\subsection{Problem Formulation}

Traditional ASAS systems typically use classification or regression models to score student answers, focusing solely on assigning a numerical score. While effective for grading, these systems lack the ability to provide detailed feedback or explain the rationale behind a score, limiting their usefulness for student learning. In addition to such numeric scores and labels, ASAS-F systems additionally generate detailed elaborated feedback \cite{shute2008focus} that explains the reasoning behind the score.

Formally, the ASAS-F system can be expressed as:

\begin{equation}
  \text{ASAS-F}(q, a, s) \rightarrow (y, l, f)
\end{equation}

where given a question $q$, a reference answer  $a$, and a student answer $s$, the system assigns:
\begin{itemize}
  \item A score $y \in [0, 1]$ (indicating correctness),
  \item A label $l \in \{\text{correct}, \text{incorrect}, \text{partially correct}\}$,
  \item Feedback $f$ that explains the reasoning behind the score.
\end{itemize}

In a zero-shot ASAS-F setting, the system utilizes the knowledge embedded in pre-trained LLMs to provide scoring and feedback without requiring labeled training data. When labeled data $\mathcal{D}$  is available, we explore two few-shot ASAS-F approaches: one using automatic prompt optimization, and another using RAG for enhanced performance.


\subsection{ASAS-F-Z: Zero-Shot ASAS-F}

\begin{figure*}[h]
  \centering 
  \includegraphics[width=1\columnwidth]{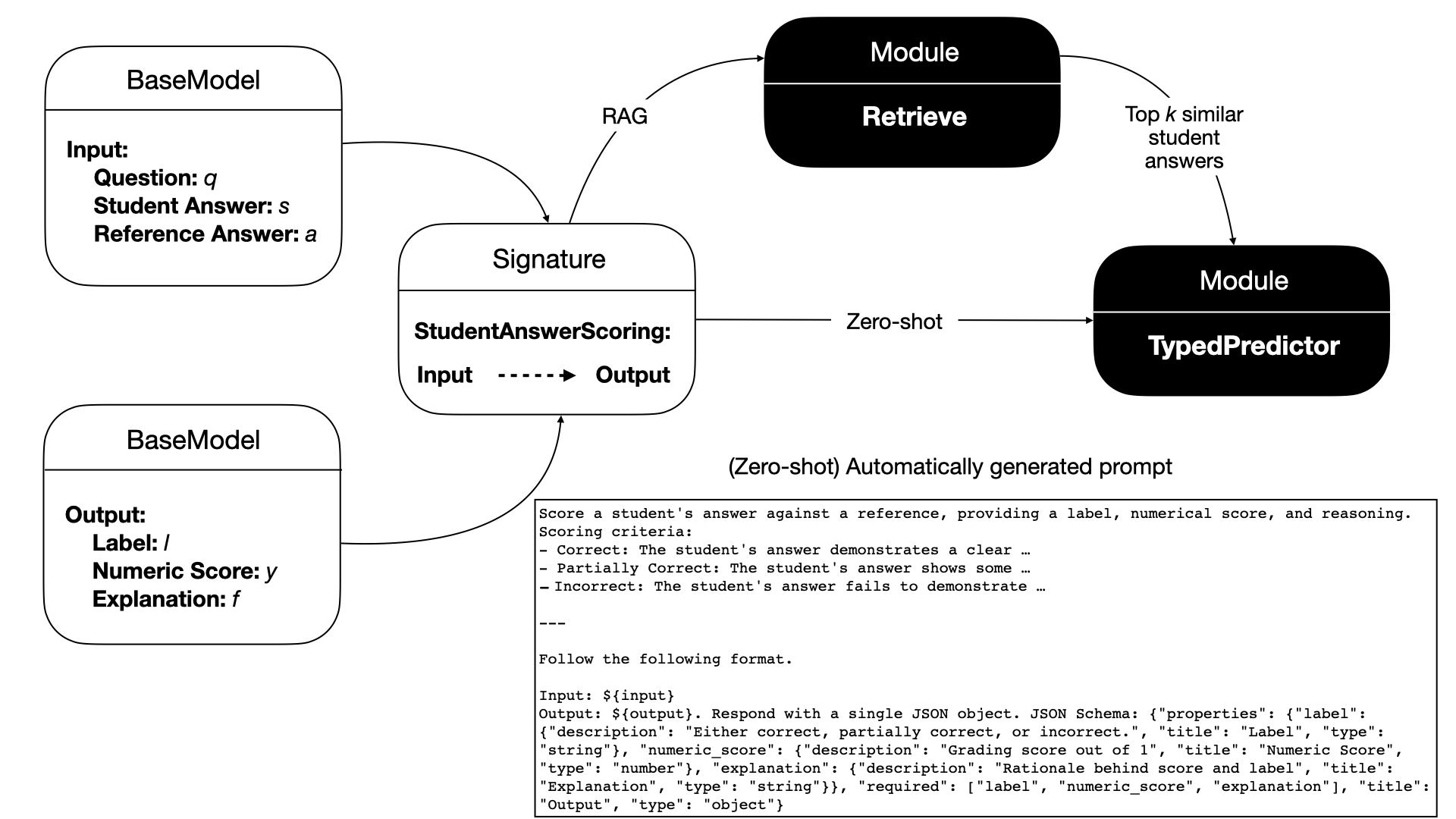} 
  \caption{Overview of the implementation of the modular ASAS-F-Z and ASAS-F-RAG systems using DSPy}  
  \label{fig:flow-diagram} 
\end{figure*}

The modular zero-shot ASAS-F approach leverages the extensive domain knowledge embedded in pre-trained LLMs to score student answers and provide feedback without requiring any labeled training data. This approach takes advantage of the generalization capabilities of LLMs, which have been trained on vast amounts of text data and possess a broad understanding of language and context.

One of the significant challenges in implementing a zero-shot ASAS-F system is effective prompt engineering. Crafting prompts that elicit the desired responses from LLMs can be complex and often involves intricate string manipulation. This process is not only time-consuming but also prone to errors.

To address these challenges, we utilize DSPy \cite{khattab2023dspy}, a framework designed to automate prompt generation and refinement to realize our ASAS-F-Z and ASAS-F-RAG systems as shown in Figure \ref{fig:flow-diagram}.

We first define our base inputs and outputs with their corresponding types and build a signature. A signature replaces a hand-written prompt by specifying what a function should do rather than how to do it. In its most basic form, a signature consists of input and output fields, each with a type. To add more control, we add a basic description of what the signature does, i.e., score a student answer and generate feedback. We also add basic scoring criteria to the signature. Instead of manually crafting prompts with different prompt engineering techniques, we use predefined modules in DSPy such as the `Predictor' module or the `Chain-Of-Thought' module that can be easily replaced or modified to generate prompts.  These modules allows us to automatically generate prompts by processing the input and output fields, generating instructions, and creating a template for the specified signature. 

In the few-shot setting, which is discussed in Section \ref{sec:asasf-rag}, an additional `Retrieve' module is added to retrieve examples from the training data. A code snippet that shows the basic implementation using DSPy to generate prompts for the zero-shot ASAS-F system is given in Appendix \ref{sec:appendix-dspy}.

\subsection{ASAS-F-Opt: Automatic Few-Shot Optimization with DSPy}

In scenarios where labeled training data is available, optimizing both prompts and few-shot examples can significantly enhance the performance of ASAS systems. This section introduces ASAS-F-Opt, which leverages DSPy for the automatic optimization of prompts and examples within the ASAS-F framework. Unlike ASAS-F-Z, where DSPy was primarily used to establish a modular system and generate prompts, ASAS-F-Opt extends this functionality. It employs a systematic optimization approach to evaluate and refine combinations of prompts and examples based on defined performance metrics.

We start by defining a performance metric for model evaluation. Then, we construct a Bayesian surrogate model that samples and assesses various combinations of prompts and examples. Our approach utilizes the MIPROv2 prompt optimization algorithm \cite{opsahl2024optimizing}, ensuring that the most effective prompts and examples are identified to enhance scoring and feedback generation.

For simplicity, we use the accuracy of the generated labels as our optimization metric. We use the Llama3:70b \cite{touvron2023llama} as the prompt generation model, and the smaller Mistral:7b \cite{jiang2024mixtral} functions as the task executor. The prompt proposal model generates initial prompts based on the source code and analyzes the dataset to produce example traces. The optimizer then systematically selects and refines these examples based on their performance according to our defined metric.  A code snippet of the implementation can be found in Appendix \ref{sec:appendix-dspy}.


\subsection{ASAS-F-RAG: Few-Shot ASAS-F Using RAG}
\label{sec:asasf-rag}

\begin{figure}[h]
  \centering 
  \includegraphics[width=\columnwidth]{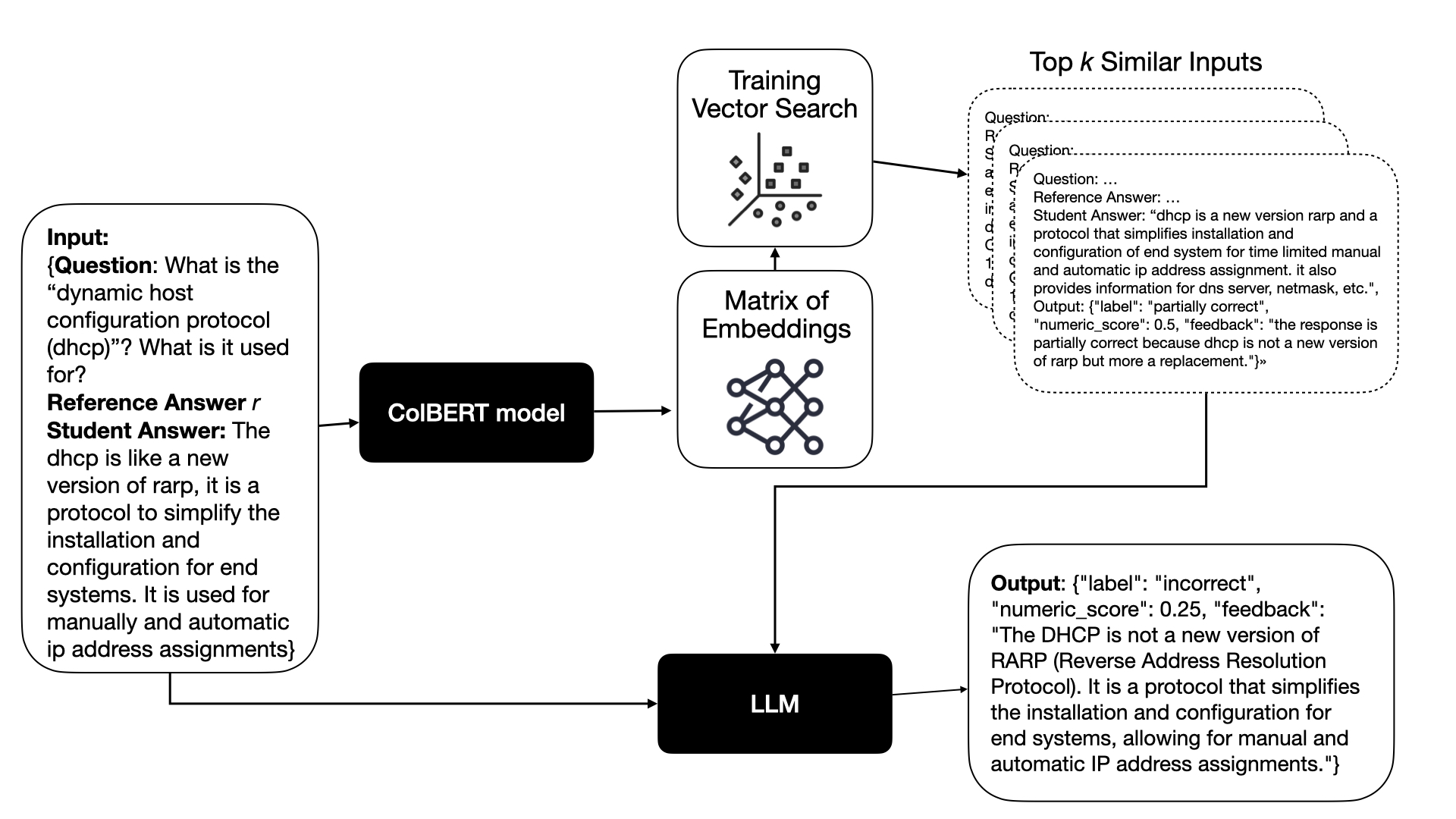} 
  \caption{Overview of the ASAS-F system using LLMs and ColBERT-driven RAG.}  
  \label{fig:diagram} 
\end{figure}

In this section, we first introduce a basic similarity-based majority-vote classifier using ColBERT for ASAS-F. We then extend this approach to incorporate RAG, which enhances the generative process by providing the LLM with contextually rich examples.

\subsubsection{Similarity-Based Majority-Vote with ColBERT}

Many existing ASAS systems use similarity-based methods to assign scores to student answers \cite{bexte2023similarity}. These systems typically compare a student’s response to reference answers or high-scoring examples, relying on similarity metrics to determine the score. 

In our few-shot ASAS-F system, we build upon this approach by not limiting the comparison to a single reference answer. Instead, we retrieve the most similar examples from the training data to enhance the prompt. We hypothesize that this method will guide the LLM to produce feedback that more closely aligns with human evaluations, resulting in more accurate feedback. 

To implement this, we utilize the ColBERT model \cite{khattab2020colbert}, which efficiently retrieves contextually relevant examples from the training data. Unlike traditional models that encode the entire input into a single vector, ColBERT encodes the input into a matrix of contextual token-level embeddings. This approach allows us to capture fine-grained similarities between the student’s answer and the training examples. Each word in the training example is represented by a BERT-based embedding, and these embeddings can be pre-stored for efficiency.

The relevance score between a student’s answer $s$  and an example $d$ is calculated using a sum of maximum similarity (MaxSim) operators between the tokens in $s$ and $d$. For each token in $s$, ColBERT identifies the most contextually similar token in $d$  and sums these similarities to compute the overall relevance score. An example is considered more relevant if it has tokens that are highly contextually similar to those in the student’s answer.

Formally, a student’s answer $s$ is tokenized, prepended with [CLS] and [Q] (query) tokens, and passed through BERT to produce vectors $[s_1, \dots, s_N]$. The example $d$ undergoes a similar process with [CLS] and [D] (document) tokens. A linear layer then adjusts the output size, and the student answer vectors and example vectors are normalized to unit length, yielding $E_s$ and $E_d$, the final vector sequences. The ColBERT score is computed as:
\begin{equation}
score(s, d) = \sum_{i=1}^{N} \max_{j=1}^{m} \left( E_{s_i} \cdot E_{d_j} \right)
\end{equation}

As a preliminary experiment, we assess the effectiveness of using the ColBERT retriever by employing a similarity-based majority-vote approach to classify student answers. This approach involves two key steps: retrieval and majority-vote classification. Initially, ColBERT encodes the student’s answer and the training examples into embeddings and retrieves the top  $k$ most similar examples based on these embeddings. Then, for each student answer,
the classifier aggregates the scores of the $k$ retrieved examples and assigns the most frequent score as the final classification and the mean score as the final numeric score.
  
\subsubsection{ASAS-F-RAG}
  
Since we focus on generating feedback, we use LLMs in combination with ColBERT to retrieve the most similar examples from the training data and ground the feedback in real examples. We define this approach as ASAS-F-RAG.
For each student answer, we retrieve the top $k = 3$ or $k = 5$ most similar examples from the training data. Each example consists of the question, the reference answer and the student answer. These examples are then fed into the LLM to generate feedback. Figure \ref{fig:diagram} shows an overview of the ASAS-F-RAG system. The ColBERT retriever is used to retrieve the most similar examples, which are then passed to the LLM to generate feedback. The feedback is then used to assign a score to the student answer. A code snippet of the implementation of the system can be found in Appendix \ref{sec:appendix-dspy}

\section{Experimental Setup}


\subsection{Dataset}
 We evaluate our system on the Short Answer Feedback (SAF) Dataset \cite{filighera2022your}. The SAF dataset consists of short answer questions in communication network topics, each with a reference answer, given score, label and conent-focused elaborated feedback. There are no other datasets, based on our review, that provide such feedback for short answers. The dataset is split into training (70\%), unseen answers (UA) (12\%) and unseen questions (UA) (18\%). The test split of UA includes new answers to the existing training questions, while the UQ split introduces novel questions. However, it is important to note that in our approach, both splits are considered novel or unseen questions, as we do not fine-tune or train the model on the training split. Even in the few-shot setting, we utilize less than 0.5\% of the training data as the provided examples.

 \subsection{Evaluation Metrics}

 \subsubsection{Scoring Metrics}

 To evaluate the performance of our ASAS-F system in terms of scoring, we used two key metrics for the classification of the generated label: accuracy and macro-averaged F1 score. These metrics assess how well the system assigns correct labels (e.g., correct, incorrect, partially correct) to student answers. For the generated numeric score, we employed Root Mean Squared Error (RMSE) to measure the difference between the predicted scores and the actual scores. These metrics provide a solid foundation for assessing the system’s ability to deliver accurate and reliable scores.

\subsubsection{Feedback Evaluation}

Evaluating the quality of generated feedback is more complex due to its subjective nature and the need to verify accuracy. We approached this evaluation using both automated metrics and human assessments.

\begin{figure}[h]
  \centering 
  \captionsetup{justification=centering}

  \includegraphics[width=0.8\columnwidth]{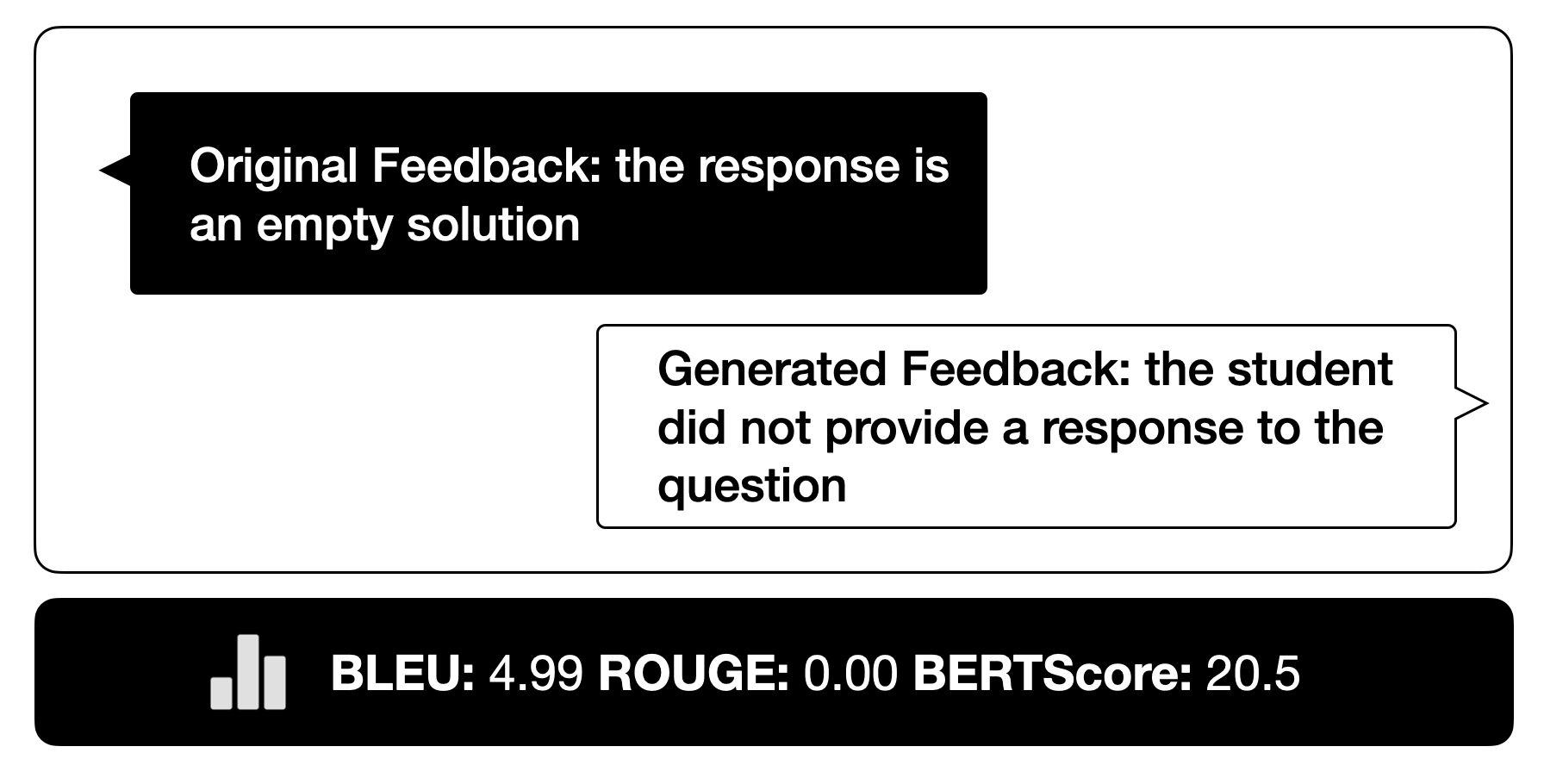} 
  \caption{Example of feedback generated by the ASAS-F system compared to the reference feedback. Traditional metrics may not capture the nuances of feedback quality.}  
  \label{fig:feedback-example} 
\end{figure}

For automatic evaluation, we utilized traditional metrics such as BLEU, ROUGE, and the more recent BERTScore. While these metrics provide a baseline for comparison,  they fail to fully capture the nuances of feedback quality. For example, in Figure \ref{fig:feedback-example}, we observe a case where the generated feedback may be semantically accurate but have low BLEU or ROUGE scores due to a lack of overlapping n-grams.
To address these limitations, we focused on two critical human-evaluated aspects of feedback to ensure a comprehensive and reliable analysis.

\begin{itemize}
  \item \textbf{Accuracy:} How well the facts in the feedback align with both the reference answer and the student’s answer.
  \item \textbf{Clarity:} How specific, clear, and coherent the feedback is.
\end{itemize}

We conducted a human evaluation using five experience teachers from the Information Technology and Networks Field employed through the Prolific platform. The annotation was conducted on a total of 108 randomly chosen samples, which included six unseen answer samples for two questions and three unseen question samples for one question per model, across 12 different models. This evaluation allowed us to ensure that the feedback provided was not only factually correct but also clear and actionable for students. The raters spent an average of seven hours on this task and were compensated at the recommended rate of £9 per hour.

\subsection{Model Selection}

 We employ various large language models, namely Mistral:7b,  Mixtral:8x22b \cite{jiang2024mixtral}, Llama3:8b, and Llama3:70b \cite{touvron2023llama} selected for their advanced performance and availability. We benchmark our ASAS-F system against several baselines: a majority class classifier, the T5 model \cite{raffel2020exploring} fine-tuned on the SAF dataset \cite{filighera2022your}, and the Llama2:7b model fine-tuned on the SAF dataset \cite{katuka2024investigating}. The Llama2:7b finetune baseline is only available for the regression task, and has not been evaluated on the label classification task.

 \section{Results}

 \subsection{Scoring Performance Analysis}
 
 \subsubsection{Similarity-Based Majority-Vote with ColBERT}
 
 \begin{table}[]
   \captionsetup{justification=centering}
   \centering
   \caption{Performance of the similarity-based majority-vote classifier using ColBERT.} 
   \label{tab:colbert-results}
   \begin{tabular}{l|ccc|ccc}
            & \multicolumn{3}{c|}{\textbf{UA Split}}                          & \multicolumn{3}{c}{\textbf{UQ Split}}                          \\ \hline
            & \textbf{Acc} & \textbf{F1} & \multicolumn{1}{l|}{\textbf{RMSE}} & \textbf{Acc} & \textbf{F1} & \multicolumn{1}{l}{\textbf{RMSE}} \\ \hline
   Majority & 0.54         & 0.234       & 0.470                              & 0.471        & 0.214       & 0.512                             \\ \hline
   k = 3    & 0.694        & 0.674       & 0.266                              & 0.602        & 0.646       & 0.277                             \\
   k = 5    & 0.739        & 0.725       & 0.259                              & 0.607        & 0.260       & 0.653                            
   \end{tabular}
 \end{table}
 
 We evaluate the performance of the retrieval-based majority-vote classification approach on accuracy, F1 and RMSE. This serves as a preliminary experiment. Table \ref{tab:colbert-results}  
 shows the performance of the ColBERT-based majority-vote classifier on the SAF dataset. The results indicate that the ColBERT-based majority-vote classifier outperforms the majority class baseline across all metrics showing the efficiency of using our refined approach.
 
 \subsubsection{ASAS-F-Z}

   \begin{table}[h]
     \caption{ASAS-F-Z results on the SAF dataset. Bold values indicate the best performance overall for each metric.}
     \label{tab:zero-shot-results}
     \centering  
     \begin{tabular}{lccc|ccc}
                                          & \multicolumn{3}{c|}{UA split}                    & \multicolumn{3}{c}{UQ split}                     \\ \hline
     \multicolumn{1}{l|}{Model}           & Acc            & F1             & RMSE           & Acc            & F1             & RMSE           \\ \hline
     \multicolumn{1}{l|}{Majority}        & 0.540          & 0.234          & 0.470          & 0.471          & 0.214          & 0.512          \\
     \multicolumn{1}{l|}{T5:finetune}     & \textbf{0.750} & \textbf{0.759} & 0.269          & 0.674          & 0.697          & 0.248          \\
     \multicolumn{1}{l|}{Llama2:finetune} & -              & -              & 0.257          & -              & -              & -              \\ \hline
     \multicolumn{1}{l|}{Mistral:7b}      & 0.643          & 0.602          & 0.283          & \textbf{0.755} & \textbf{0.751} & 0.212          \\
     \multicolumn{1}{l|}{Llama3:8b}       & 0.448          & 0.476          & 0.335          & 0.495          & 0.489          & 0.290          \\
     \multicolumn{1}{l|}{Mixtral:8x22b}   & 0.726          & 0.706          & 0.263          & 0.695          & 0.708          & 0.213          \\
     \multicolumn{1}{l|}{Llama3:70b}      & 0.691          & 0.678          & \textbf{0.253} & 0.716          & 0.742          & \textbf{0.199}
     \end{tabular}
     \end{table}

\begin{figure*}[h]
   \centering
   \subfloat[][Unseen Answers]{
     \label{fig:zero-shot-ua}
 \includegraphics[width=0.5\columnwidth]{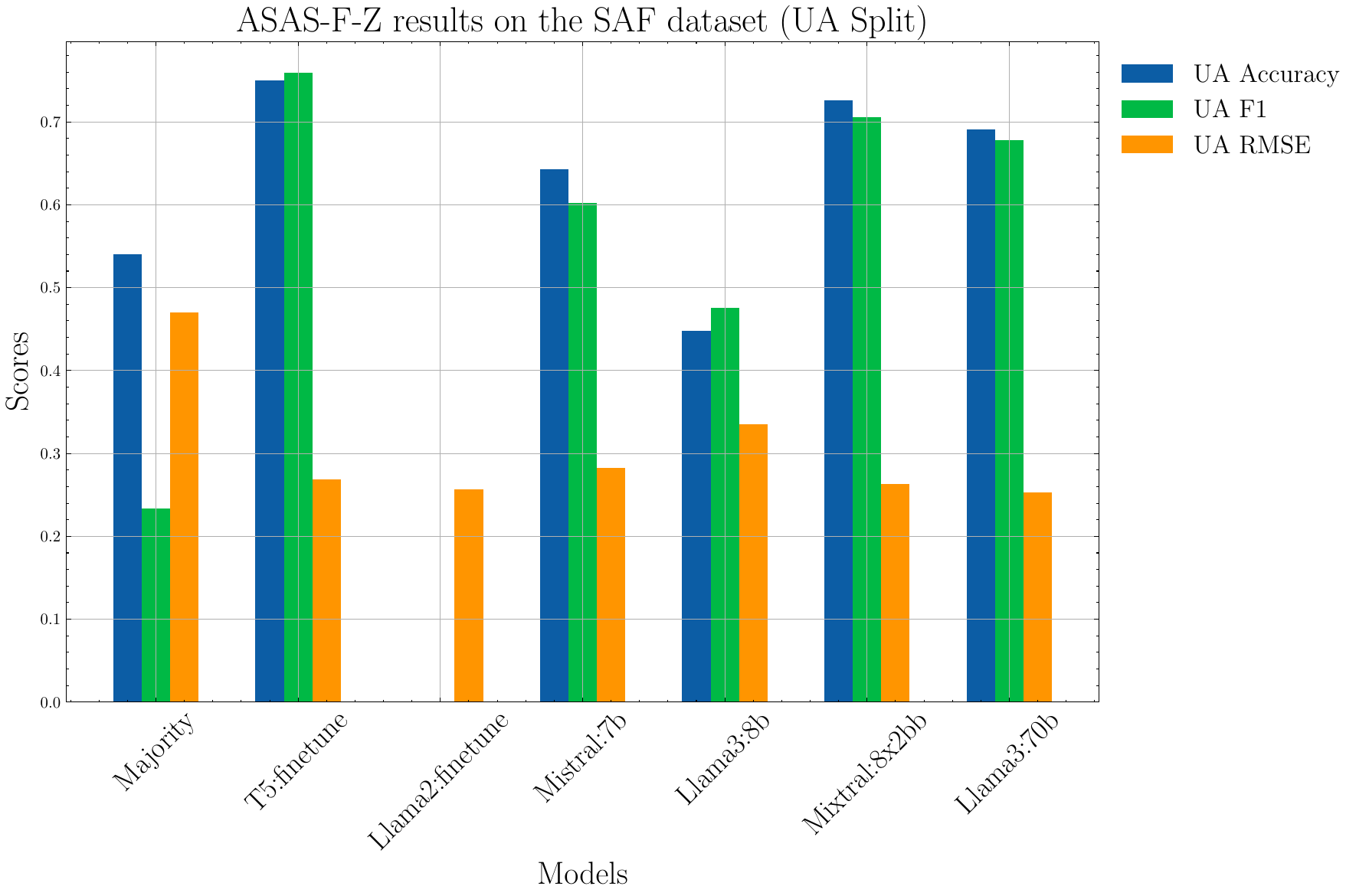}}
   \subfloat[][Unseen Questions]{
     \label{fig:zero-shot-uq}
     \includegraphics[width=0.5\columnwidth]{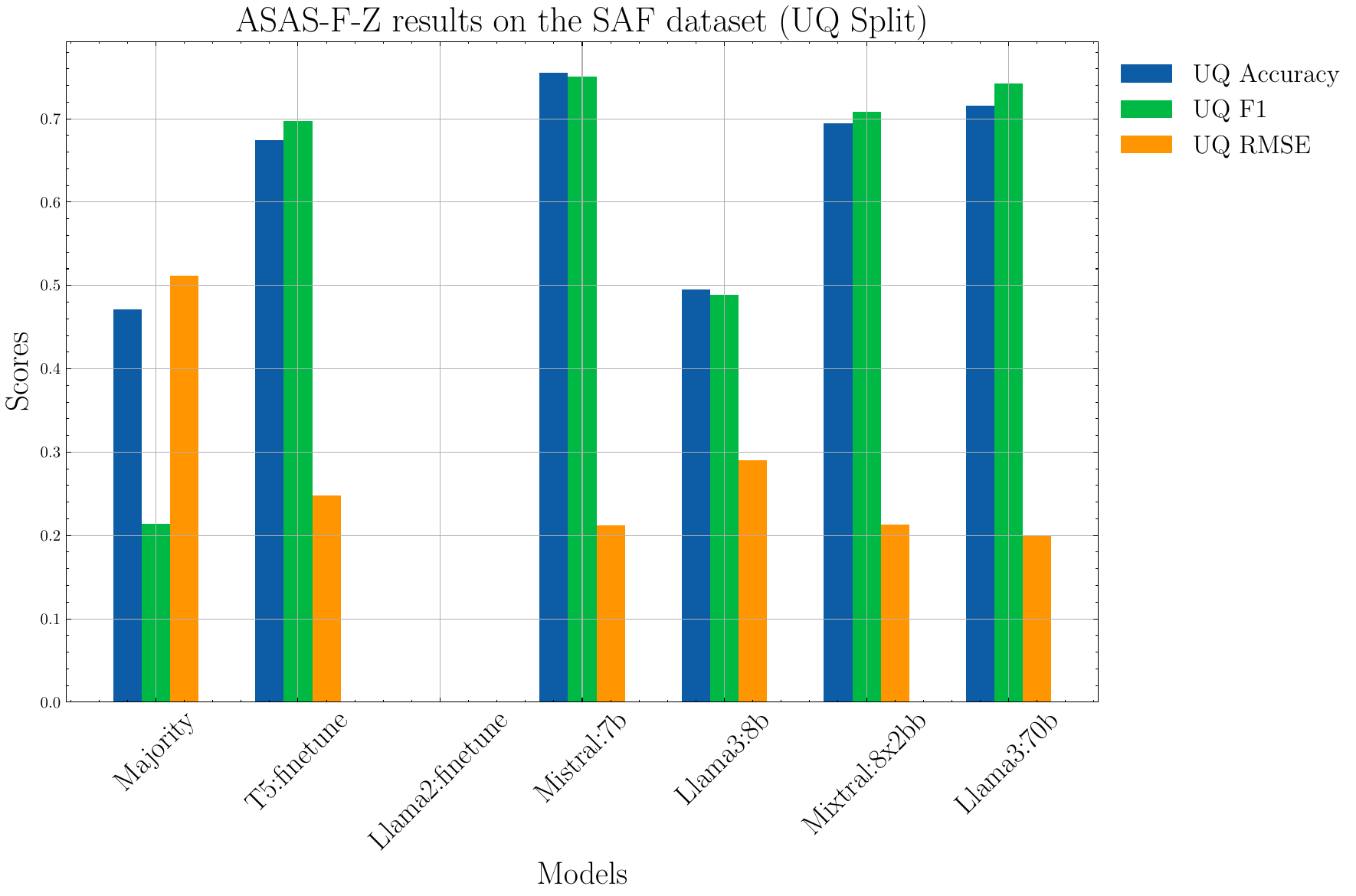}}
   \captionsetup{justification=centering}
   \caption{Performance of the ASAS-F-Z system on the SAF dataset. Higher is better for accuracy and F1 score, lower is better for RMSE.}
   \label{figure:zero-shot}
 \end{figure*}

   Table \ref{tab:zero-shot-results} and Figures \ref{fig:zero-shot-ua} and \ref{fig:zero-shot-uq} show
 the results of the zero-shot ASAS-F system on both test splits of the SAF dataset compared to the baselines.  In the UA split, the fine-tuned baselines outperform the zero-shot system in terms of accuracy and F1 score. However, in the UQ split, excluding the Llama3:8b model, the ASAS-F-Z system is able to outperform all baselines in all metrics. The highest performing model is the smallest Mistral:7b model, which achieves an accuracy of 0.755 and an F1 score of 0.751 in the UQ split. Llama3:70b achieves the lowest RMSE. This suggests that while finetuning can perform similarly to human raters in answers that it has seen during training, it tends to underperform when evaluating on new questions. The zero-shot ASAS-F system, on the other hand, is able to generalize better to new questions, indicating the potential of this approach for educational applications.

 \subsubsection{ASAS-F-Opt}

 In ASAS-F-Opt, we aim to enhance the model’s performance by refining the selection and utilization of examples. While the ASAS-F-RAG system dynamically retrieves relevant examples to inform responses, ASAS-F-Opt focuses on optimizing how these examples are integrated into the model's processing. This approach seeks to ensure that the selected few-shot examples are maximally effective in improving accuracy. We use the bigger Llama3:70b model to refine the prompts and few-shot examples, and utilize the  Mistral:7b for generating the outputs during testing since it was able to outperform in ASAS-F-Z. 
 
 The results for the UA split were below expectations, with accuracy of 0.627, F1 Score of 0.41, and RMSE of 0.309. Similarly, in the UQ split, the metrics were accuracy of 0.643, F1 Score of 0.668, and RMSE of 0.317. These results were notably inferior to the performance of ASAS-F-Z using Mistral:7b. This outcome suggests that the automatic few-shot optimization method may not be effective in enhancing the model’s performance in this particular scenario. The use of pre-defined, specified few-shot examples may not always be optimal, indicating that automatic optimization strategies require further refinement. This highlights the complexity of automatic optimization and the need for a more nuanced approach to designing these strategies to better align with the requirements of the ASAS-F system.
 
 \subsubsection{ASAS-F-RAG}
 
 \begin{table}[h]
   \centering
   \caption{ASAS-F-RAG results on the SAF dataset. Bold values indicate the best performance overall for each metric.}
   \label{tab:few-shot-results}
   \resizebox{\columnwidth}{!}{%
   \begin{tabular}{l|ccc|ccc}
                        & \multicolumn{3}{c|}{UA split}   & \multicolumn{3}{c}{UQ split}   \\ \hline
   Model                & Accuracy & F1 Score & RMSE & Accuracy & F1 Score & RMSE \\ \hline
   Majority             & 0.540    & 0.234    & 0.470 & 0.471    & 0.214    & 0.512 \\
   T5:finetune          & 0.750    & 0.759    & 0.269 & 0.674    & 0.697    & 0.248 \\
   Llama2:finetune      & -        & -        & 0.257 & -        & -        & -     \\ \hline
   k = 3, Mistral:7b    & 0.710    & 0.684    & 0.232 & 0.753    & 0.761    & 0.230 \\
   k = 5, Mistral:7b    & 0.710    & 0.667    & 0.240 & \textbf{0.760} & \textbf{0.768} & 0.209 \\
   k = 3, Llama3:8b     & 0.663    & 0.672    & 0.259 & 0.628    & 0.664    & 0.238 \\
   k = 5, Llama3:8b     & 0.742    & 0.739    & 0.271 & 0.581    & 0.621    & 0.217 \\
   k = 3, Mixtral:8x22b & 0.679    & 0.616    & 0.230 & 0.693    & 0.619    & \textbf{0.204} \\
   k = 5, Mixtral:8x22b & 0.702    & 0.638    & 0.212 & 0.677    & 0.626    & 0.215 \\
   k = 3, Llama3:70b    & \textbf{0.766} & 0.775    & 0.213 & 0.693    & 0.726    & 0.220 \\
   k = 5, Llama3:70b    & \textbf{0.766} & \textbf{0.778} & \textbf{0.208} & 0.714    & 0.749    & 0.213 \\
   \end{tabular}
    }
 \end{table}

 \begin{figure}[]
   \centering
   \subfloat[Unseen Answers]{
     \label{fig:few-shot-ua}
 \includegraphics[width=0.5\columnwidth]{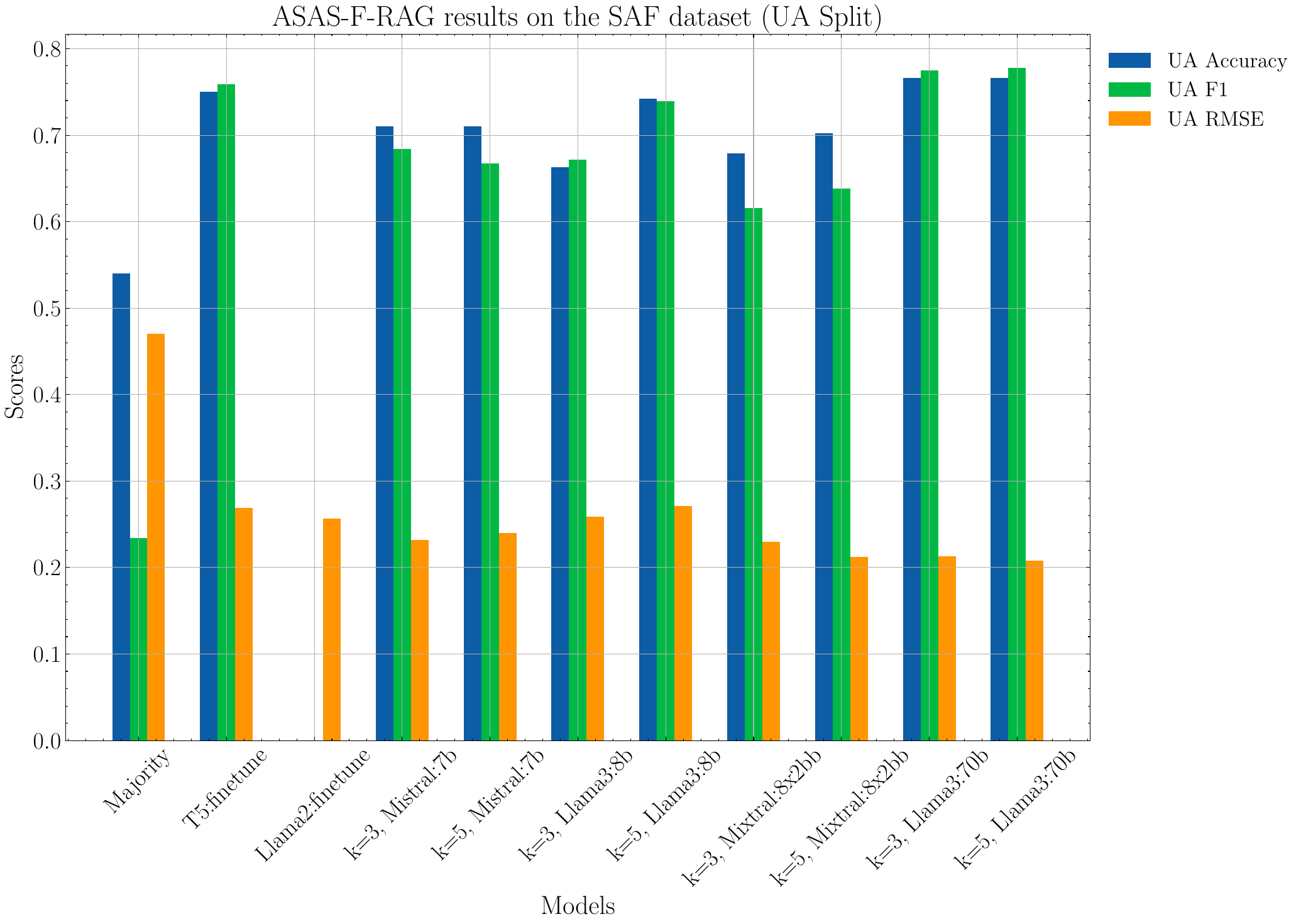}}
   \subfloat[Unseen Questions]{
     \label{fig:few-shot-uq}
     \includegraphics[width=0.5\columnwidth]{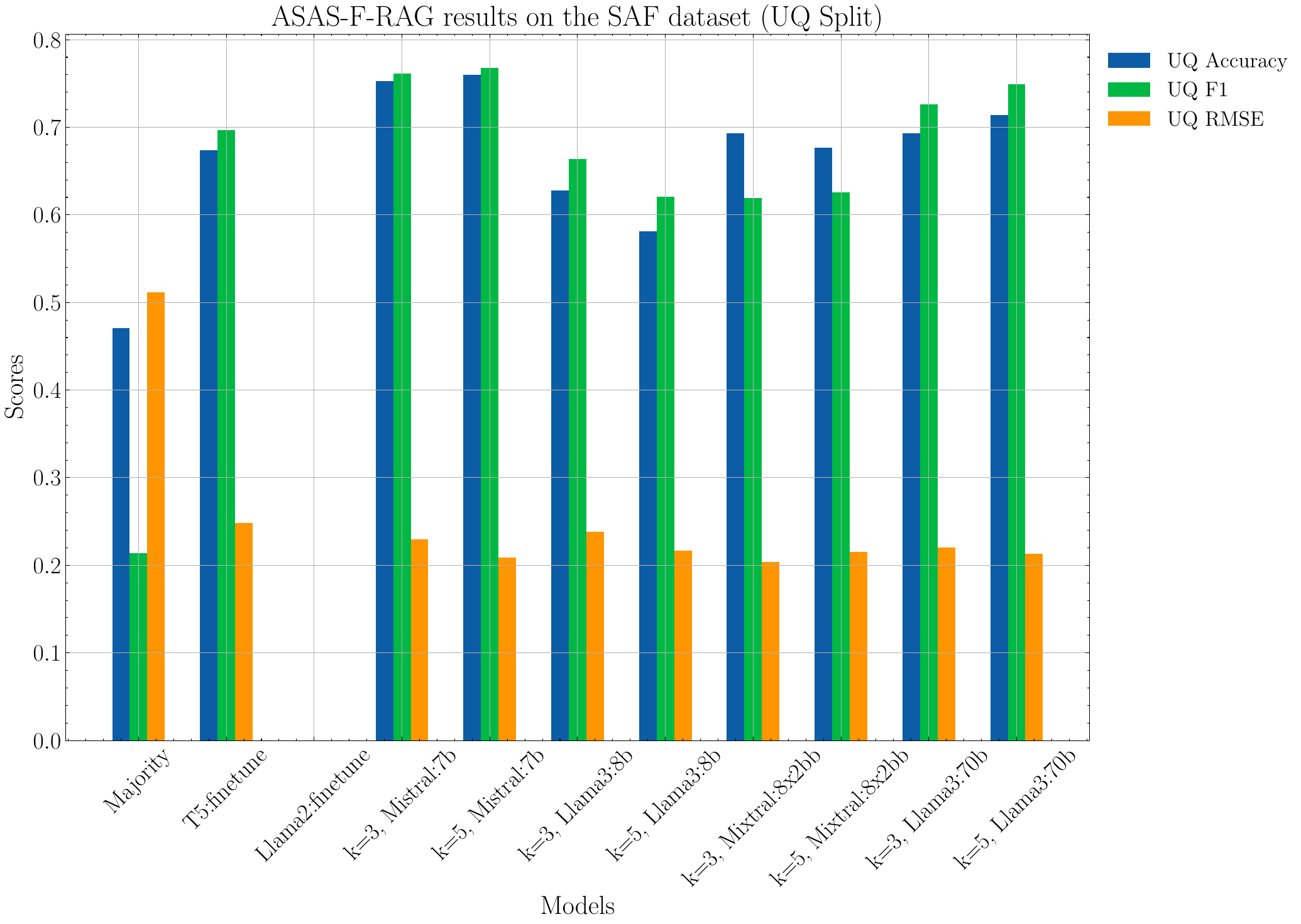}}
   \captionsetup{justification=centering}
   \caption{Performance of the ASAS-F-RAG system on the SAF dataset. Higher is better for accuracy and F1 score, lower is better for RMSE.}
   \label{figure:few-shot}
 \end{figure}
 
 To address the limitations shown by the ASAS-F-Z system especially in the UA split and the low performance shown by ASAS-F-Opt, we explore the ASAS-F-RAG system. This system focuses on automatically retrieving the most similar answers from existing data to augment the prompts used to improve model performance.
 Table \ref{tab:few-shot-results} shows the results of the ASAS-F-RAG system on both test splits of the SAF dataset compared to the baselines. Figures \ref{fig:few-shot-ua} and \ref{fig:few-shot-uq} illustrate the performance of the ASAS-F-RAG system on the UA  and UA splits, respectively.
 
 In the UA split, the results highlight that the Llama3:70b model with five examples consistently achieved the highest performance across all metrics, demonstrating superior accuracy, F1 score, and RMSE compared to other models. This indicates that larger models with more examples can leverage their extensive training to provide more accurate predictions in the few-shot setting. The other models also exhibited a similar trend.
 
 Conversely, in the UQ split, the performance improvements with additional examples are less consistent. Notably, the Llama3:8b and Mixtral:8x22b models did not follow the general trend of improved performance with more examples. The Llama3:70b model, while still competitive, did not maintain its leading position as it did in the UA split. Here, the zero-shot Llama3:70b model actually outperformed the few-shot models, suggesting that the examples used in the few-shot setting might not be as effective when dealing with questions that differ significantly. 
 
 The discrepancies in performance between the two splits can be attributed to the nature of the retrieved examples. In the unseen questions split, the examples retrieved were from different questions, which may have introduced variability that impacted the model’s ability to generalize effectively. This variability might explain why the performance with more examples did not consistently improve and why the zero-shot model performed comparably or better in some cases.

 \subsection{Feedback Quality Analysis}
 
 \subsubsection{Statistical Analysis}
 
 Table \ref{tab:feedback-evaluation-stats} presents the evaluation of feedback quality on the SAF dataset using SacreBLEU \cite{post2018call}, ROUGE-2 \cite{lin-2004-rouge}, and BERTScore \cite{zhang2019bertscore} metrics. It is evident that the baseline models were able to outperform both the zero-shot and few-shot ASAS-F systems in all metrics. However, the baseline models due to finetuning have been reported to often copy common phrases from the training data. Looking at the second-best performing models, we can see that the ASAS-F-RAG outperforms the ASAS-F-Z in both splits. This suggests that incorporating a small number of labeled examples can improve the quality of the generated feedback in terms of consistency with the reference feedback. However, increasing the number of examples beyond a certain threshold does not necessarily lead to higher similarity to the reference feedback. This can be seen in the UQ split where the second-best performing models used 3 examples.

 \begin{table}[]
   \centering
   \caption{Statistical feedback quality evaluation on the SAF dataset. Bold values indicate the best performance overall for each metric. Underline values indicate the second-best performance.}
 
   \resizebox{\columnwidth}{!}{%
   \begin{tabular}{c|ccc|ccc}
                        & \multicolumn{3}{c|}{UA split}                                   & \multicolumn{3}{c}{UQ split}                  \\ \hline
   Model                & BLEU                & ROUGE               & BERTScore           & BLEU          & ROUGE         & BERTScore     \\ \hline
   Majority             & 2.2                 & 20.2                & 42.2                & 0.2           & 11.5          & 38.1          \\
   T5:finetune          & 33.7                & \textbf{52.8}       & \textbf{65.0}         & \textbf{10.7} & \textbf{31.1} & \textbf{52.2} \\
   Llama2:finetune      & \textbf{39.6}       & 13.7                & -                   & -             & -             & -             \\ \hline
   k = 0, Mistral:7b    & 2.42                & 5.11                & 23.6                & 2.15          & 6.93          & 25.9          \\
   k = 3, Mistral:7b    & 7.90                & 23.7                & 40.2                & 6.38          & {\underline{28.8}}    & {\underline{ 43.9}}    \\
   k = 5, Mistral:7b    & 6.75                & 21.7                & 37.7                & 3.82          & 19.2          & 33.6          \\
   k = 0, Llama3:8b     & 1.76                & 3.97                & 17.5                & 1.45          & 4.06          & 18.6          \\
   k = 3, Llama3:8b     & 9.86                & 28.5                & 43.3                & 4.97          & 24.7          & 39.3          \\
   k = 5, Llama3:8b     & 11.9                & 32.5                & 45.6                & 5.36          & 24.7          & 39.7          \\
   k = 0, Mixtral:8x22b & 2.42                & 26.7                & 25.6                & 2.72          & 6.20          & 28.0          \\
   k = 3, Mixtral:8x22b & 13.4                & 29.2                & 42.4                & {\underline{ 8.73}}    & 23.6          & 35.0          \\
   k = 5, Mixtral:8x22b & 11.6                & 26.7                & 41.0                & 3.78          & 17.6          & 30.4          \\
   k = 0, Llama3: 70b   & 1.81                & 4.11                & 20.5                & 2.10          & 5.35          & 21.6          \\
   k = 3, Llama3:70b    & 13.2       & 34.7                & 49.5                & 3.67          & 19.3          & 35.7          \\
   k = 5, Llama3:70b    & {\underline{15.0}} & {\underline{36.3}} & {\underline{50.6}} & 3.86          & 20.1          & 36.0         
   \end{tabular}
   }
   \label{tab:feedback-evaluation-stats}
   \end{table}
 
 \subsubsection{Human Evaluation}
 
 \begin{figure*}[h]
   \centering
   \subfloat[][Accuracy]{
     \label{fig:human-accuracy-evaluation}
 \includegraphics[width=0.5\columnwidth]{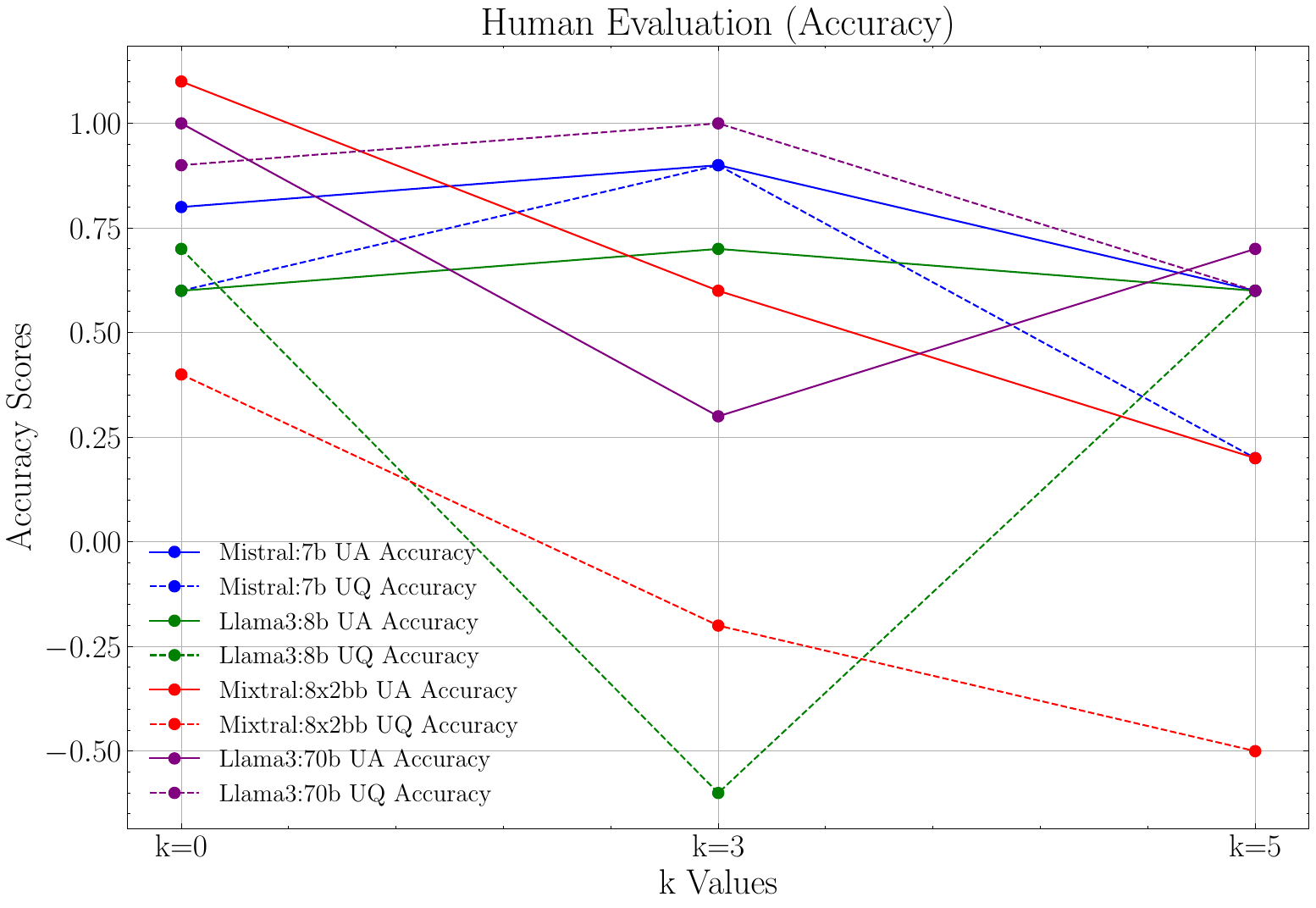}}
   \subfloat[][Clarity]{
     \label{fig:human-clarity-evaluation}
     \includegraphics[width=0.5\columnwidth]{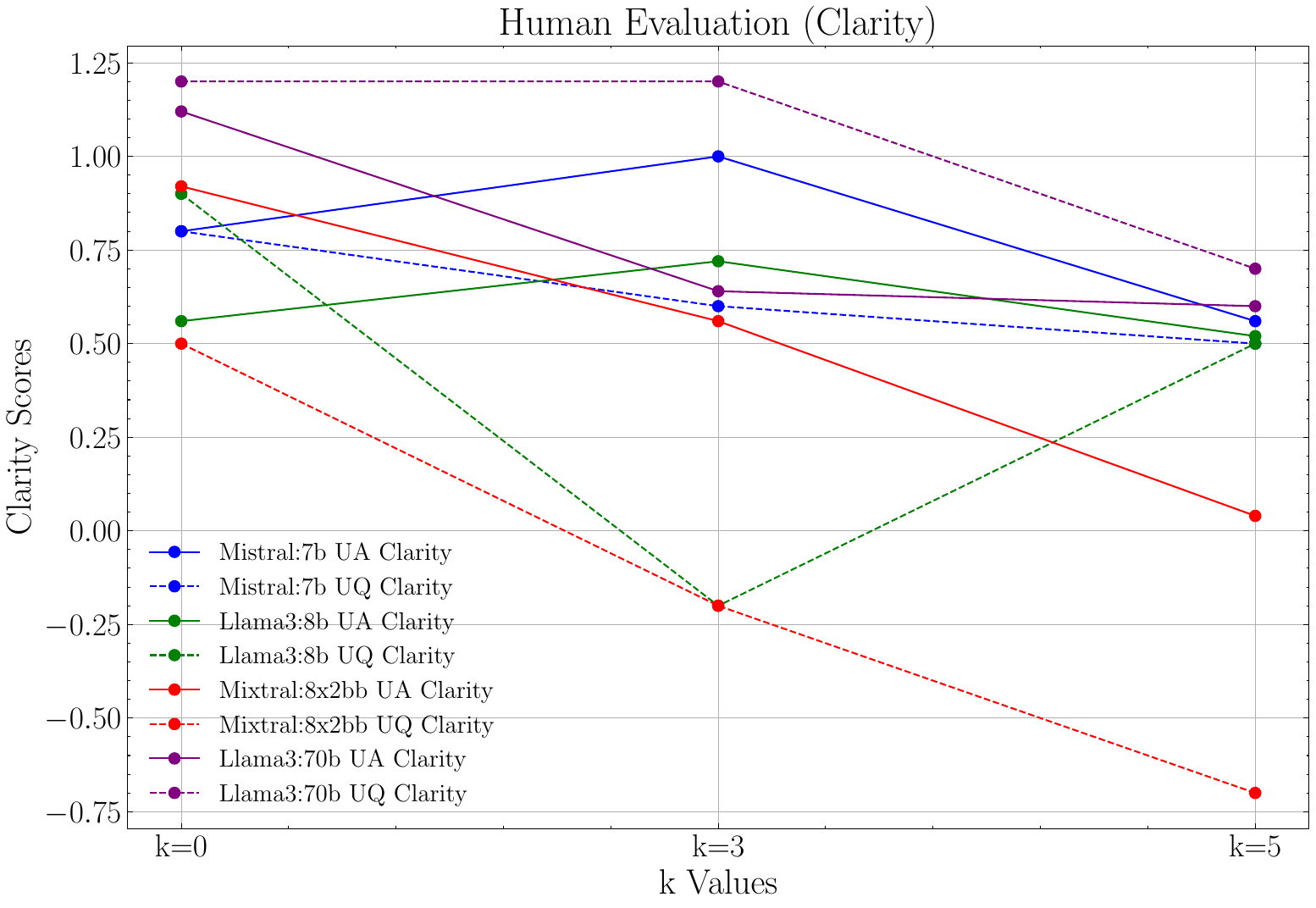}}
   \captionsetup{justification=centering}
   \caption{Human feedback evaluation results on samples with existing student answers. Higher is better.}
   \label{figure:human-evaluation}
 \end{figure*}

 \begin{table}[]
   \centering
   \caption{Human feedback quality evaluation results on the samples with existing student answers from the SAF dataset. Bold values indicate the best performance overall for each metric while underline values indicate the second-best performance.}
   \label{tab:human-evaluation}
   \begin{tabular}{l|cc|cc}
                            & \multicolumn{2}{c|}{\textbf{UA Split}} & \multicolumn{2}{c}{\textbf{UQ Split}} \\ \hline
                            & \textbf{Accuracy}  & \textbf{Clarity}  & \textbf{Accuracy}  & \textbf{Clarity} \\ \hline
   k = 0, Mistral:7b        & 0.84               & 0.80              & 0.60               & 0.80             \\
   k = 3, Mistral:7b        & 0.92               & {\underline{ 1.00}}              & {\underline{ 0.90}}               & 0.60             \\
   k = 5, Mistral:7b        & 0.60               & 0.56              & 0.20               & 0.50             \\ \hline
   \textit{Average}         & 0.79               & 0.79              & 0.57               & 0.63             \\ \hline
   k = 0, Llama3:8b         & 0.64               & 0.56              & 0.70               & {\underline{ 0.90}}             \\
   k = 3, Llama3:8b         & 0.68               & 0.72              & -0.60              & -0.20            \\
   k = 5, Llama3:8b         & 0.40               & 0.52              & 0.60               & 0.50             \\ \hline
   \textit{Average}         & 0.57               & 0.60              & 0.23               & 0.40             \\ \hline
   k = 0, Mixtral:8x22b     & \textbf{1.12}               & 0.92              & 0.40               & 0.50             \\
   k = 3, Mixtral:8x22b     & 0.56               & 0.56              & -0.20              & -0.20            \\
   k = 5, Mixtral:8x22b     & 0.16               & 0.04              & -0.50              & -0.70            \\ \hline
   \textit{Average}         & 0.61               & 0.51              & -0.10              & -0.13            \\ \hline
   k = 0, Llama3:70b        & {\underline{ 1.04}}               & \textbf{1.12}              & {\underline{ 0.90}}               & \textbf{1.20}             \\
   k = 3, Llama3:70b        & 0.32               & 0.64              & \textbf{1.00}               & \textbf{1.20}             \\
   k = 5, Llama3:70b        & 0.68               & 0.60              & 0.60               & 0.70             \\ \hline
   \textit{Average}         & 0.68               & 0.79              & 0.83               & 1.03             \\ \hline
   \textit{k = 0, Average}  & 0.91               & 0.85              & 0.65               & 0.85             \\
   \textit{k = 3, Average}  & 0.62               & 0.73              & 0.28               & 0.35             \\
   \textit{k = 5, Average}  & 0.46               & 0.43              & 0.23               & 0.25             \\ \hline
   \textit{Overall Average} & 0.66               & 0.67              & 0.35               & 0.48            
   \end{tabular}
   \end{table}

     To ensure the quality of the feedback generated by our system, we conducted a human evaluation with five expert teachers in Information Technology and Networks. The raters assessed 108 randomly selected samples from 12 models, focusing on accuracy and clarity. Feedback was rated on a 5-point Likert scale ranging from -2 to 2, where -2 indicated strongly inaccurate or unclear feedback, 0 indicated neutral feedback, and 2 indicated strongly accurate or clear feedback. Figures \ref{fig:human-accuracy-evaluation} and \ref{fig:human-clarity-evaluation} illustrate the human evaluation results for samples with existing student answers. Table \ref{tab:human-evaluation} presents the overall results, excluding no-response samples. 

      For the baseline models, human evaluation was not conducted. Table \ref{tab:feedback-comparison} provides a sample of the feedback generated by the baseline model and samples from our ASAS-F system. The feedback generated by the baseline model is repetitive and lacks the depth needed for comprehensive understanding. A high degree of similarity could artificially inflate the statistical evaluation metrics like BLEU scores. The high overlap between training inputs and generated outputs signals that the model may not truly be learning to adapt to novel inputs but is instead reproducing familiar patterns. Thus, while the baseline system might achieve high scores on automated metrics due to repetition of training data, its semantic contribution may be minimal compared to the more informative and contextually appropriate feedback generated by our system. More examples of comparisons between the feedback generated by the baseline models and our ASAS-F system can be found in Appendix \ref{sec:appendix-output}.

  \begin{table}[]
    \centering
    \caption{Comparison of feedback generated by the baseline models and samples from the ASAS-F system. Underlined text are repetitions or redundancies.}
    \label{tab:feedback-comparison}
    \begin{tabular}{l}
    \hline
    \textbf{Student Answer 1}                                                                                                                                                                                                                                                                                                                                                                                                                                                                                                                                            \\ \hline
    \begin{tabular}[c]{@{}l@{}}The extension headers are placed between fixed header and payload. \\ The main advantage is that they are optional. This means that the use of \\ space is more efficient because there won’t be any redundant information \\ placed there in order to keep the expected length.\end{tabular}                                                                                                                                                                                                                                                  \\ \hline
    \textbf{Gold Feedback}                                                                                                                                                                                                                                                                                                                                                                                                                                                                                                                                               \\ \hline
    \begin{tabular}[c]{@{}l@{}}The response answers only the location of extension headers correctly. \\ The definition is missing and the advantage is incorrect when compared to \\ IPv4 as the option field in IPv4 is also optional. Thus, there is no\\ efficiency benefit.\end{tabular}                                                                                                                                                                                                                                                                                \\ \hline
    \textbf{T5 Generated Feedback}                                                                                                                                                                                                                                                                                                                                                                                                                                                                                                                                       \\ \hline
    \begin{tabular}[c]{@{}l@{}}the advantage is incorrect as the option field in the ipv4 header is optional \\ and can be placed between payload and fixed header. \underline{the advantage is} \\ \underline{incorrect as the option field in the ipv4 header is optional} \\ and can be placed between payload and fixed header.\end{tabular}                                                                                                                                                                                                                                                      \\ \hline
    \textbf{k = 5, Llama3:70b Generated Feedback}                                                                                                                                                                                                                                                                                                                                                                                                                                                                                                                        \\ \hline
    \begin{tabular}[c]{@{}l@{}}the response correctly states the location of extension headers between \\ the fixed header and payload. however, the main advantage stated is \\ not entirely accurate. while it’s true that ipv6 extension headers are \\ optional, this doesn’t directly contribute to more efficient use of space. \\ a correct advantage would be that they allow for adding new options \\ without changing the fixed header size or that ipv6 packets with optional \\ headers are typically processed faster/simpler by intermediate devices.\end{tabular} \\ \hline
    \textbf{k = 3, Mistral:7b Generated Feedback}                                                                                                                                                                                                                                                                                                                                                                                                                                                                                                                        \\ \hline
    \begin{tabular}[c]{@{}l@{}}The response correctly identifies the location of extension headers and \\ mentions that they are optional. However, it does not fully address the \\ main advantage as stated in the reference answer, which includes both the \\ ability to append new options without changing the header and the \\ simplified processing by intermediate devices.\end{tabular}                                                                                                                                                                             
    \end{tabular}
\end{table}

     In the analysis, we find that inconsistencies across different models occasionally confused raters, leading to lower accuracy scores. Notably, for no-response answers, which comprised 22\% of all samples, large language models (LLMs) were generally able to generate feedback that correctly identified the lack of an answer. In 23 out of 24 cases, the feedback was considered strongly accurate. In the remaining case, the feedback failed to explicitly state the absence of an answer. However, disagreement among the raters was observed in all cases regarding the accuracy and clarity. Despite this, the inter-rater agreement for these no-response samples was relatively high, with a Krippendorff’s alpha of 0.77 for accuracy and 0.71 for clarity. \footnote{Krippendorff's alpha is a statistical measure of agreement among raters, accounting for the possibility of chance agreement. It ranges from 0 to 1, with values closer to 1 indicating strong agreement.}
 
     In contrast, for the remaining 78\% of samples—where student answers were present—Krippendorff’s alpha dropped significantly to 0.19 for accuracy and 0.20 for clarity, indicating low agreement. This discrepancy may stem from the granularity of the 5-point scale and the inherent subjectivity of evaluating nuanced feedback.
 
     Given the confusion that occured with the no-responses, we focus on the samples with existing student answers for the human evaluation. Among the models evaluated, the Llama3:70b model achieved the highest average accuracy and clarity scores in both the UA and UQ splits. The Mistral:7b model followed closely, showing competitive performance. However, a general trend of decreasing accuracy and clarity with an increasing number of examples was observed across all models. 
 
     As a post hoc analysis, we merged the strongly accurate and accurate categories and the strongly inaccurate and inaccurate categories to create a 3-point scale. This adjustment led to an increase in the Krippendorff’s alpha to 0.91 for accuracy and 0.89 for clarity for the no-response answers. For the remaining answers, the Krippendorff’s alpha increased to 0.60 for accuracy and 0.64 for clarity, indicating a moderate level of agreement among raters. This suggests that the original scale was too granular and that a simpler scale may have resulted in higher agreement scores.

     \begin{figure}[h]
       \centering 
       \includegraphics[width=\columnwidth]{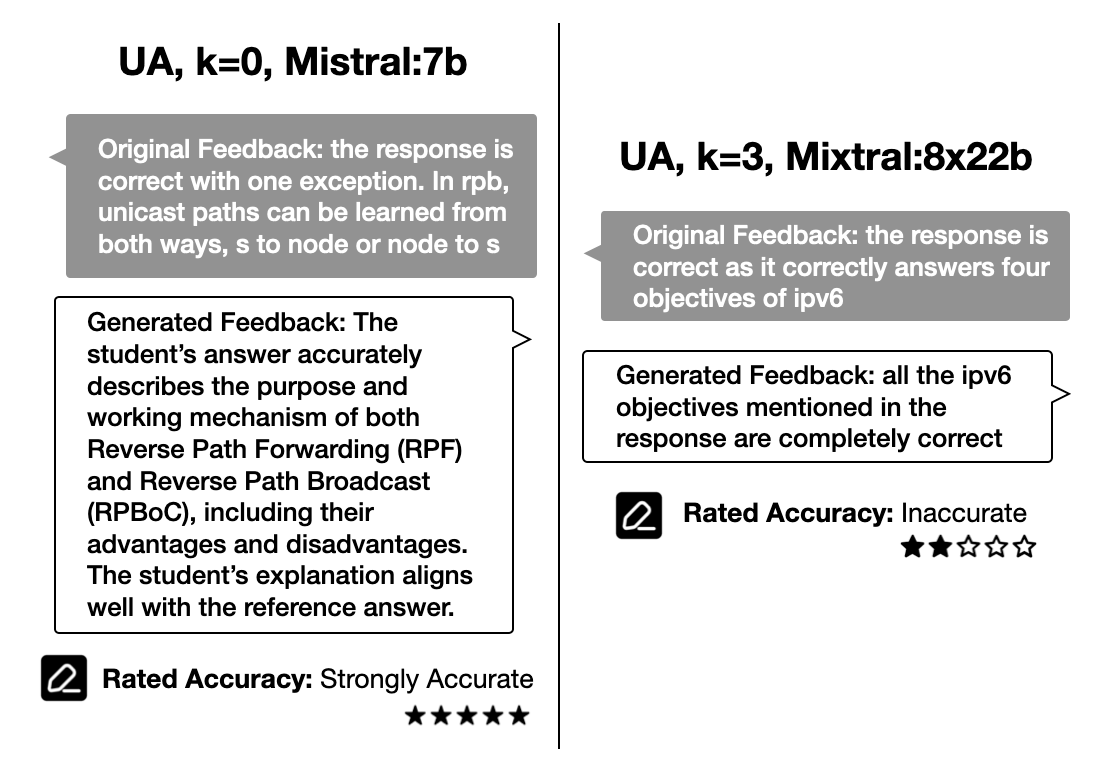} 
       \caption{Samples of human evaluation accuracy ratings for generated feedback vs actual feedback.}  
       \label{fig:ratings} 
     \end{figure}

     In the few-shot approach, LLMs often mimicked the style of the provided reference answers or examples as shown in the first example in Figure \ref{fig:ratings}. Many raters commented the feedback that was too brief, lacking sufficient context or explanation. Although the feedback was factually correct, it did not provide the depth needed for comprehensive understanding, which could be critical in educational settings. Conversely, in the zero-shot approach, the generated feedback was often longer and appeared more detailed. However, as the second example in Figure \ref{fig:ratings} demonstrates, while some of this feedback was labeled as accurate, it was actually incorrect when compared to the reference answers. This highlights the challenge of calibrating LLMs, as they may present incorrect feedback with high confidence \cite{zhou2024relying}, making it difficult to detect when they are hallucinating \cite{ji2023survey}, especially when evaluating the feedback in isolation.

   \section{Discussion}

   In this study, we aimed to answer key research questions regarding the performance of an ASAS-F system in zero-shot and few-shot settings. Rather than relying on intricate and labor-intensive prompt engineering to optimize performance for specific datasets, we adopted a modular design for our system using DSPy. We explored two main methods for selecting few-shot examples ASAS-F-RAG and ASAS-F-Opt in comparison with the simpler ASAS-F-Z and fine-tuned baselines. Our evaluation focused on the system’s ability to score student answers that align with the reference answers and generate feedback that is accurate and clear.

   To address \textbf{RQ1} (How does the performance of our modular ASAS-F system compare to state-of-the-art models in automatic short answer scoring?), our findings revealed that LLMs can compete with fine-tuned baselines in scoring accuracy. While fine-tuning on unseen answers showed better performance compared to the zero-shot approach, three out of four LLMs surpassed the fine-tuned baselines when tested on unseen questions, with the smallest model, Mistral:7b, delivering the best results. Our few-shot approach consistently outperformed fine-tuned baselines in both evaluation splits, underscoring the effectiveness of RAG for improving scoring performance in ASAS tasks. Our modular system and the LLMs generally performed well in terms of the evaluation metrics. However, when enforcing type constraints on the outputs, some LLMs failed to abide by them. For these output errors, we implemented a fallback mechanism to ensure comprehensive evaluation (see Appendix \ref{sec:appendix-errors}).
   
   For \textbf{RQ2} (When labeled training data is available, how can we optimize prompts and few-shot examples to improve our ASAS-F performance in an efficient way?), we experimented with a Bayesian optimization approach to automatically optimize the selection of few-shot examples and prompts. However, we found that our dynamic few-shot approach using RAG significantly outperformed the automatic optimization method. In tasks like ASAS, where the subjectivity of scoring plays a critical role, dynamic selection of similarly scored examples proves more effective than fully automated methods. This indicates that aligning retrieved examples closely with the scoring task’s subjectivity is crucial for optimizing ASAS performance.
   
   Regarding \textbf{RQ3} (How accurate and clear is the feedback generated by our ASAS-F system?), traditional evaluation metrics like BLEU and ROUGE are often insufficient, as they do not account for the nuanced nature of educational feedback. These metrics can fail to capture the quality of feedback when generated sentences are factually correct but differ significantly from the reference text. In our human evaluation, we observed considerable disagreement among raters on the accuracy and clarity of the feedback generated by the LLMs, reflecting the inherent subjectivity of the task. Although the few-shot approach generated feedback more aligned with reference feedback, its accuracy ratings were lower than those from the zero-shot approach. This highlights the challenges in evaluating feedback quality, particularly in subjective tasks. Future work should explore more robust evaluation frameworks for feedback generation, focusing on both subjective and objective measures.

   \section{Conclusion}

   This study provides a comprehensive evaluation of an ASAS-F system in both zero-shot and few-shot settings. We introduced a novel RAG approach, leveraging ColBERT for automatic short answer scoring with feedback (ASAS-F). Our results demonstrated that the few-shot approach not only outperformed the more computationally expensive fine-tuned baselines in scoring accuracy but also performed effectively in zero-shot settings, especially on unseen questions, highlighting the robustness of our method for automatic scoring.
   
   However, when evaluating feedback quality, we observed an inverse relationship between the number of examples provided and the quality of the feedback generated, as assessed by human experts. This suggests that while our system excels in scoring, generating high-quality, pedagogically sound feedback remains challenging, particularly due to the subjective nature of feedback evaluation. Additionally, the inherent complexity of feedback generation in educational settings, combined with issues like model calibration and potential hallucinations, highlights the difficulty of evaluating feedback produced by large language models.
   
   Despite these challenges, our approach—combining open-source LLMs with powerful retrieval models in a modular, computationally efficient framework—shows strong potential to enhance the accuracy and efficiency of ASAS systems. Future research should focus on improving feedback quality, exploring model calibration techniques, and developing more reliable methods for evaluating feedback in educational contexts.

\section*{Acknowledgements}
This work was partially supported by JST (Japan Science and Technology Agency), the establishment of university fellowships towards the creation of science technology innovation, Grant Number JPMJFS2132, and JSPS (Japan Society for the Promotion of Science)  KAKENHI No. JP21H00907 and JP23H03511.

\bibliographystyle{unsrtnat} 
\bibliography{custom}

 \appendix

\section{DSPy Code Snippets}
\label{sec:appendix-dspy}

\begin{figure}[]
  \centering 
  \includegraphics[width=\columnwidth]{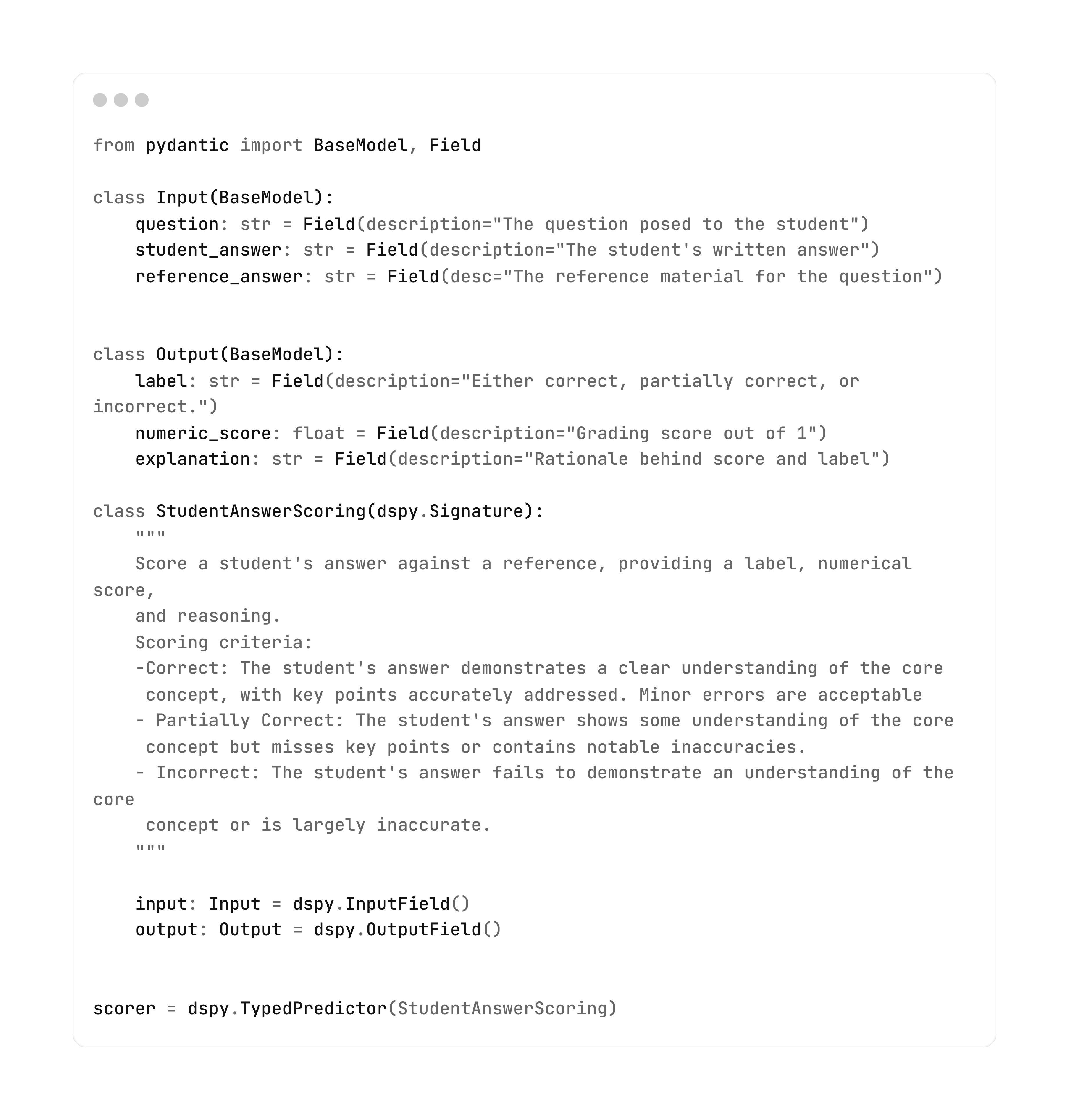} 
  \caption{DSPy code snippet of the ASAS-F-Z system.}  
  \label{fig:dspy-zero-shot} 
\end{figure}

\begin{figure}[]
  \centering 
  \includegraphics[width=\columnwidth]{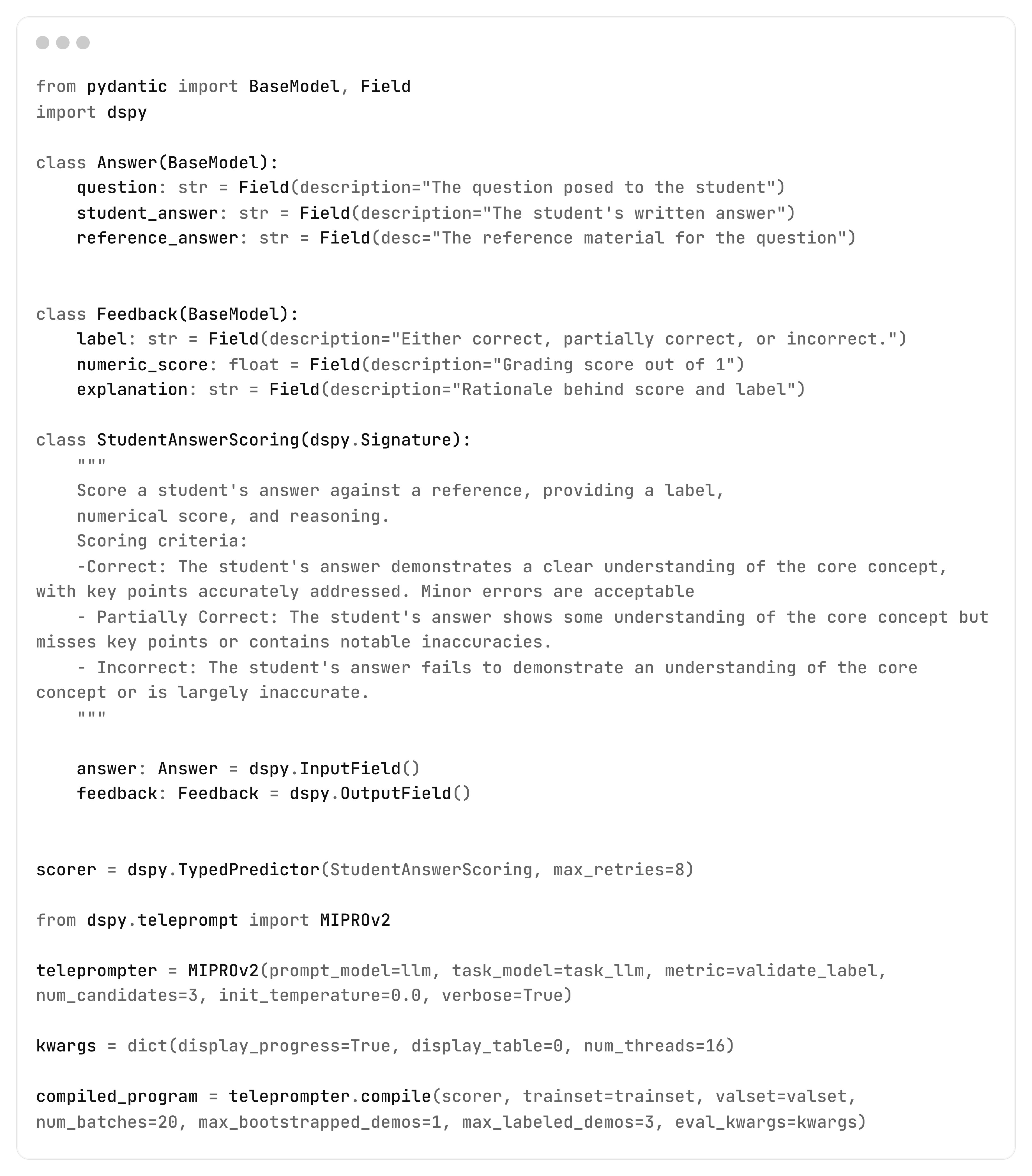} 
  \caption{Code snippet of the ASAS-F-Opt system.}
  \label{fig:dspy-opt} 
\end{figure}

\begin{figure}[]
  \centering 
  \includegraphics[width=\columnwidth]{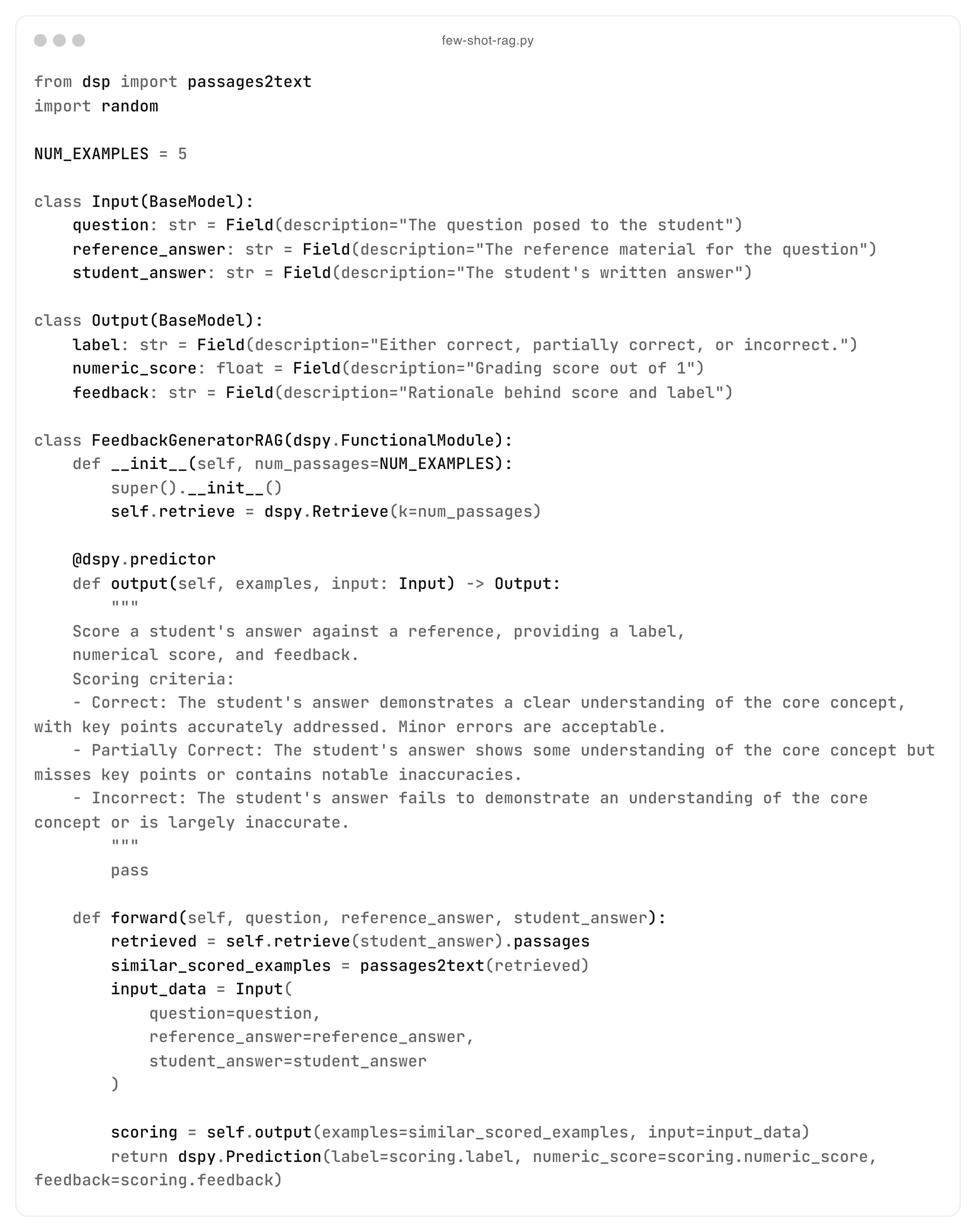} 
  \caption{DSPy code snippet of the ASAS-F-RAG system.}  
  \label{fig:dspy-rag} 
\end{figure}

In this section, we provide code snippets of the DSPy implementation of the ASAS-F system. The code snippets for ASAS-F-Z, ASAS-F-Opt and ASAS-F-RAG are shown in Figures \ref{fig:dspy-zero-shot}, \ref{fig:dspy-opt}, and \ref{fig:dspy-rag}, respectively.

\section{Output Generation Examples}
\label{sec:appendix-output}

This section provides a comparison of feedback samples generated by the baseline models and our system. Tables \ref{tab:feedback-comparison-2}, \ref{tab:feedback-comparison-3}, and \ref{tab:feedback-comparison-4} present feedback generated by the baseline models and samples from our system for the same answers. These samples show clear instances of phrase reuse from the fine-tuning process from the baseline, highlighting limited generalization capabilities compared to the more informative and contextually appropriate feedback generated by our system.

\begin{table}[]
  \centering
  \caption{Comparison of feedback generated by the baseline models and samples from the ASAS-F system. Underlined text are repetitions or redundancies.}
  \label{tab:feedback-comparison-2}
  \resizebox{\columnwidth}{!}{%
  \begin{tabular}{l}
  \hline
  \textbf{Student Answer 2}                                                                                                                                                                                                                                                                                                                                                                                                                                                                                                                                                                                                         \\ \hline
  \begin{tabular}[c]{@{}l@{}}1. IPv6 extension headers contain supplementary information used by network \\ devices (such as routers, switches, and endpoint hosts) to decide how \\ to direct or process an IPv6 packet. 2. They are located between fixed \\ header and payload. 3. IPv6 extension headers can be of arbitrary length. \\ The number of options that a packet carries is not limited.\end{tabular}                                                                                                                                                                                                                  \\ \hline
  \textbf{Gold Feedback}                                                                                                                                                                                                                                                                                                                                                                                                                                                                                                                                                                                                            \\ \hline
  \begin{tabular}[c]{@{}l@{}}The response answers the description, location of extension headers, and \\ advantage correctly.\end{tabular}                                                                                                                                                                                                                                                                                                                                                                                                                                                                                           \\ \hline
  \textbf{T5 Generated Feedback}                                                                                                                                                                                                                                                                                                                                                                                                                                                                                                                                                                                                    \\ \hline
  \begin{tabular}[c]{@{}l@{}}\underline{the advantage is partially correct} as it does not state what extension \\ headers are, how they are used in ipv4, and how they are used in \\ ipv6. \underline{the advantage is incorrect} as the option field in the ipv4 \\ header is optional and can be of arbitrary length.\end{tabular}                                                                                                                                                                                                                                                                                                                         \\ \hline
  \textbf{k = 0, Llama3:8b, Predicted Label, Score \& Feedback}                                                                                                                                                                                                                                                                                                                                                                                                                                                                                                                                                                     \\ \hline
  \begin{tabular}[c]{@{}l@{}}The student’s answer demonstrates a good understanding of the concept of \\ IPv6 extension headers, correctly stating that they contain supplementary \\ information and are located between the fixed header and payload. However, \\ their explanation is not entirely accurate, as they mention arbitrary length \\ without specifying the limitations on the number of options carried by a \\ packet. Additionally, while they touch upon the main advantage of extension \\ headers compared to IPv4, their answer lacks clarity and does not fully address \\ the reference’s correct points.\end{tabular} \\ \hline
  \textbf{k = 5, Mixtral8:22xb, Predicted Label, Score \& Feedback}                                                                                                                                                                                                                                                                                                                                                                                                                                                                                                                                                                 \\ \hline
  \begin{tabular}[c]{@{}l@{}}the response answers the description and location of the extension \\ headers correctly but does not mention any advantage compared to ipv4.\end{tabular}                                                                                                                                                                                                                                                                                                                                                                                                                                               
  \end{tabular}
  }
\end{table}

\begin{table}[]
  \centering
  \caption{Comparison of feedback generated by the baseline models and samples from the ASAS-F system. Underlined text indicates repetitions or redundancies.}
  \label{tab:feedback-comparison-3}
  \resizebox{\columnwidth}{!}{%
  \begin{tabular}{l}
  \hline
  \textbf{Student Answer 3}                                                                                                                                                                                                                                                                                                                                                                                                                                                                                                                     \\ \hline
  \begin{tabular}[c]{@{}l@{}}Spanning trees are appealing to broad- and multicasting scenarios, because \\ they allow the packets to only travel one path (except travelling backwards). \\ This removes the need for looking up specific tables as in RPF/RPB.\end{tabular}                                                                                                                                                                                                                                                                      \\ \hline
  \textbf{Gold Feedback}                                                                                                                                                                                                                                                                                                                                                                                                                                                                                                                        \\ \hline
  \begin{tabular}[c]{@{}l@{}}It is true that there is a unique path between nodes but that not only does away \\ with the need to look at routing tables in RPF/RPB but reduces duplicates by \\ removing loops (unnecessary links). No explanation was provided for \\ modifying the link-state algorithm to construct a multicast spanning tree \\ for nodes.\end{tabular}                                                                                                                                                                 \\ \hline
  \textbf{T5 Generated Feedback}                                                                                                                                                                                                                                                                                                                                                                                                                                                                                                                \\ \hline
  \begin{tabular}[c]{@{}l@{}}The explanation behind using a spanning tree for multicast and broadcast is \\ \underline{partially correct as it does not mention} how the link-state algorithm is \\ modified to construct a multicast spanning tree. The explanation for modifying \\ the link-state algorithm to construct a multicast spanning tree is \\ \underline{partially} \underline { correct as it does not mention how} each node shares its multicast \\ information with others by adding it to the link-state packet.\end{tabular}                                             \\ \hline
  \textbf{k = 5, Llama3:70b Generated Feedback}                                                                                                                                                                                                                                                                                                                                                                                                                                                                                                 \\ \hline
  \begin{tabular}[c]{@{}l@{}}The response correctly answers why a spanning-tree usage is ideal in multicast \\ and broadcast scenarios, mentioning that it removes the need for looking up  \\ specific tables as in RPF/RPB. However, the description lacks detail on how \\ link-state routing can be modified to construct a spanning tree for multicasting.\end{tabular}
 
\end{tabular}
}
\end{table}

\begin{table}[htbp]
  \centering
  \caption{Comparison of feedback generated for Student Answer 4.}
  \label{tab:feedback-comparison-4}
  \resizebox{\columnwidth}{!}{%
  \begin{tabular}{l}
  \hline
  \textbf{Student Answer 4}                                                                                                                                                                                                                                                                                                                                                                                                                                                                                                                                                                                                   \\ \hline
  PROPERTY: The spanning tree does not have any loops.                                                                                                                                                                                                                                                                                                                                                                                                                                                                                                                                                                        \\ \hline
  \textbf{Gold Feedback}                                                                                                                                                                                                                                                                                                                                                                                                                                                                                                                                                                                                      \\ \hline
  \begin{tabular}[c]{@{}l@{}}The response correctly answers why a spanning-tree usage is ideal in multicast\\ and broadcast. However, the explanation for modifying the link-state algorithm\\ to construct a multicast spanning tree for nodes is not provided.\end{tabular}                                                                                                                                                                                                                                                                                                                                                          \\ \hline
  \textbf{T5 Generated Feedback}                                                                                                                                                                                                                                                                                                                                                                                                                                                                                                                                                                                              \\ \hline
  \begin{tabular}[c]{@{}l@{}}The response correctly identifies the spanning-tree property that makes it\\ appealing for broadcast and multicast and explains how the link-state\\ algorithm is modified to construct a multicast spanning tree. However,\\ the explanation for modifying the link-state algorithm is partially correct as it\\ does not state how this modification is achieved.\end{tabular}                                                                                                                                              \\ \hline
  \textbf{k = 0, Llama3:8b, Predicted Label, Score \& Feedback}                                                                                                                                                                                                                                                                                                                                                                                                                                                                                                                                                               \\ \hline
  \begin{tabular}[c]{@{}l@{}}The student correctly identifies that spanning trees do not have loops,\\ which is a key property. However, their explanation is incomplete and\\ does not fully capture the essence of why spanning trees are appealing for\\ multicasting. The reference answer provides a more comprehensive explanation,\\ including the concept of unique paths between every pair of nodes in the tree.\\ Additionally, the student’s answer does not address how to modify link-state\\ routing to construct a spanning tree for multicasting, which is an important\\ aspect of the question.\end{tabular} \\ \hline
  \textbf{k = 5, Mixtral8:22xb, Predicted Label, Score \& Feedback}                                                                                                                                                                                                                                                                                                                                                                                                                                                                                                                                                           \\ \hline
  \begin{tabular}[c]{@{}l@{}}The response only states one property of a spanning tree but fails to explain\\ how link-state routing can be modified for constructing a multicast spanning tree.\end{tabular}                                                                                                                                                                                                                                                                                                                                                                                                              
  \end{tabular}
  }
\end{table}

\section{Typed Predictor Error Analysis}
\label{sec:appendix-errors}

  In our implementation of the ASAS-F system, we primarily relied on typed predictors generated through the DSPy framework. Typed predictors are advantageous because they enforce type constraints on both input and output, ensuring that the model adheres to the predefined structure and can handle complex, multi-layered data. This structure enhances the reliability and consistency of the model’s predictions.
  
  However, during the evaluation phase, we observed an average of 4.13\% errors across all models due to formatting issues that led to no output being generated by the typed predictors. These errors arise from strict type constraints, where slight deviations in input formatting—such as unexpected characters or variations in text structure—caused the predictor to fail.
  
  To address this issue and ensure comprehensive evaluation, we implemented a fallback mechanism using a normal predictor without typed constraints. In cases where the typed predictor failed to generate an output, the normal predictor was used to obtain the predictions. This approach allowed us to maintain the flow of predictions and provided a means to evaluate the system’s performance despite the formatting errors.
  
  Table \ref{tab:typed-predictor-errors} presents the error rates for each model. Although typed predictors adhere to strict standards, some errors are unavoidable due to the complexity of the input data and the constraints on the output structure. On average, the Llama models exhibited lower error rates compared to the Mistral models, suggesting that the Llama models were better at adhering to the predefined type constraints. Among the Llama models, the smaller Llama3:8b achieved the lowest error rate at 2.04\%, followed by the Llama3:70b with a rate of 3.0\%. In contrast, the Mistral:7b model had an error rate of 4.16\%, while the Mixtral:8x22b model recorded the highest average error rate at 7.32\%.

  \begin{table}[]
    \centering
    \caption{Error Rates (\%) in Typed Predictors Across Models. Values in bold show the overall average error rate for each model.}

    \begin{tabular}{l|cc|}
    \multicolumn{1}{c|}{}                                  & \multicolumn{1}{c|}{\textbf{UA Split}} & \textbf{UQ split} \\ \hline
    k = 0, Mistral:7b                                      & \multicolumn{1}{c|}{5.16}              & 2.86              \\
    k = 3, Mistral:7b                                      & \multicolumn{1}{c|}{3.57}              & 2.86              \\
    k = 5, Mistral:7b                                      & \multicolumn{1}{c|}{6.35}              & 4.17              \\ \hline
    \multicolumn{1}{c|}{\multirow{2}{*}{\textit{Average}}} & \multicolumn{1}{c|}{5.03}              & 3.30              \\ \cline{2-3} 
    \multicolumn{1}{c|}{}                                  & \multicolumn{2}{c|}{\textbf{4.16}}                         \\ \hline
    k = 0, Llama3:8b                                       & \multicolumn{1}{c|}{3.97}              & 1.82              \\
    k = 3, Llama3:8b                                       & \multicolumn{1}{c|}{2.78}              & 0.26              \\
    k = 5, Llama3:8b                                       & \multicolumn{1}{c|}{1.59}              & 1.83              \\ \hline
    \multicolumn{1}{c|}{\multirow{2}{*}{\textit{Average}}} & \multicolumn{1}{c|}{2.78}              & 1.30              \\ \cline{2-3} 
    \multicolumn{1}{c|}{}                                  & \multicolumn{2}{c|}{\textbf{2.04}}                         \\ \hline
    k = 0, Mixtral:8x22b                                   & \multicolumn{1}{c|}{8.33}              & 6.25              \\
    k = 3, Mixtral:8x22b                                   & \multicolumn{1}{c|}{3.57}              & 11.20             \\
    k = 5, Mixtral:8x22b                                   & \multicolumn{1}{c|}{5.95}              & 8.59              \\ \hline
    \multicolumn{1}{c|}{\multirow{2}{*}{\textit{Average}}} & \multicolumn{1}{c|}{5.95}              & 8.68              \\ \cline{2-3} 
    \multicolumn{1}{c|}{}                                  & \multicolumn{2}{c|}{\textbf{7.32}}                         \\ \hline
    k = 0, Llama3:70b                                      & \multicolumn{1}{c|}{1.98}              & 0.00              \\
    k = 3, Llama3:70b                                      & \multicolumn{1}{c|}{3.97}              & 3.39              \\
    k = 5, Llama3:70b                                      & \multicolumn{1}{c|}{3.17}              & 5.47              \\ \hline
    \multicolumn{1}{c|}{\multirow{2}{*}{\textit{Average}}} & \multicolumn{1}{c|}{3.04}              & 2.95              \\ \cline{2-3} 
    \multicolumn{1}{c|}{}                                  & \multicolumn{2}{c|}{\textbf{3.0}}                         
    \end{tabular}
    \label{tab:typed-predictor-errors}
    \end{table}

\end{document}